\documentclass[letterpaper]{article} % DO NOT CHANGE THIS
\usepackage{aaai25}  % DO NOT CHANGE THIS
\usepackage{times}  % DO NOT CHANGE THIS
\usepackage{helvet}  % DO NOT CHANGE THIS
\usepackage{courier}  % DO NOT CHANGE THIS
\usepackage[hyphens]{url}  % DO NOT CHANGE THIS
\usepackage{graphicx} % DO NOT CHANGE THIS
\usepackage{caption} % DO NOT CHANGE THIS AND DO NOT ADD ANY OPTIONS TO IT
\usepackage{algorithm}
\usepackage{algorithmic}

\usepackage{caption}
\usepackage{graphicx, subfig}
\usepackage{amsmath}
\usepackage{newfloat}
\usepackage{listings}
\DeclareCaptionStyle{ruled}{labelfont=normalfont,labelsep=colon,strut=off} % DO NOT CHANGE THIS
\floatstyle{ruled}
\newfloat{listing}{tb}{lst}{}
\floatname{listing}{Listing}
\title{RepControlNet: ControlNet Reparameterization}

\author {
    \normalsize
     Zhaoli Deng\textsuperscript{\rm 1},
     Kaibin Zhou\textsuperscript{\rm 1,2},
     Fanyi Wang\textsuperscript{\rm 1}\thanks{Corresponding author},
     Zhenpeng Mi\textsuperscript{\rm 1}\thanks{Corresponding author}
}
\affiliations {
    \textsuperscript{\rm 1}Honor Device Co., Ltd,\textsuperscript{\rm 2}Tongji University\\
   {dengzhaoli, wangfanyi, mizhenpeng}@honor.com, kb824999404@tongji.edu.cn\\
}

\usepackage{bibentry}
\begin{document}

\maketitle

\begin{abstract}
% 随着扩散模型的广泛应用，高昂的推理资源消耗成为其普遍应用的重要瓶颈，可控生成作为扩散模型的研究重点方向之一，推理加速和模型压缩相关的研究更显得重要。针对这一问题，本文提出了一种模态重参数化方法RepControlNet，在不增加计算量的前提下实现扩散模型的可控生成。RepControlNet在训练过程中使用adapter将模态信息调制到特征空间，复制一份原扩散模型的CNN和MLP可学习层作为模态网络，并基于原始权重和超参系数初始化这部分权重。训练过程只优化模态网络部分参数。推理过程将模态网络中和原扩散模型权重进行模态重参数化，在不增加参数量的前提下，可以实现比拟甚至超越比如ControlNet这种使用额外参数和计算量的方法。我们在SD1.5和SDXL上均进行了大量实验，实验结果表明本文提出的RepControlNet的有效性和高效性。
With the wide application of diffusion model, the high cost of inference resources has became an important bottleneck for its universal application. Controllable generation, such as ControlNet, is one of the key research directions of diffusion model, and the research related to inference acceleration and model compression is more important. In order to solve this problem, this paper proposes a modal reparameterization method, RepControlNet, to realize the controllable generation of diffusion models without increasing  computation. In the training process, RepControlNet uses the adapter to modulate the modal information into the feature space, copy the CNN and MLP learnable layers of the original diffusion model as the modal network, and initialize these weights based on the original weights and coefficients. The training process only optimizes the parameters of the modal network. In the inference process, the weights of the neutralization original diffusion model in the modal network are reparameterized, which can be compared with or even surpass the methods such as ControlNet, which use additional parameters and computational quantities, without increasing the number of parameters. We have carried out a large number of experiments on both SD1.5 and SDXL, and the experimental results show the effectiveness and efficiency of the proposed RepControlNet.

\end{abstract}

% Uncomment the following to link to your code, datasets, an extended version or similar.
%
% \begin{links}
%     \link{Code}{https://aaai.org/example/code}
%     \link{Datasets}{https://aaai.org/example/datasets}
%     \link{Extended version}{https://aaai.org/example/extended-version}
% \end{links}

\section{Introduction}
% 随着扩散模型应用的日趋广泛和用户群体的指数化增长,扩散模型的小型化[]和推理加速研究[]变得十分必要，为实际落地应用节约推理时间和能源损耗，降低碳排放为环保事业做出无声的贡献。以snapfution[]为代表的扩散模型小型化工作，牺牲一定生成质量的前提下，通过减少模型参数量和部署蒸馏实现推理速度的提升。扩散模型的推理加速的出发点一般是进行步数蒸馏[UFO。。。]实现推理步数的压缩。以往方法都是针对扩散模型自身优化，在结合多模态信息引导的可控生成领域，相关的加速研究很少。

With the increasingly widespread application of diffusion models and the exponential growth of user groups, the research on model compression \cite{li2024snapfusion} and inference acceleration \cite{xu2024ufogen} of diffusion models have become essential. These efforts aim to save inference time and energy consumption in practical applications, thereby reducing carbon emissions and making a silent contribution to environmental protection. The model compression work represented by SnapFusion \cite{li2024snapfusion} improves inference speed by reducing the number of model parameters and deploying distillation, at the cost of sacrificing some generation quality. The goal of inference acceleration for diffusion models is generally to achieve step distillation \cite{xu2024ufogen}, thereby compressing the number of inference steps. Previous methods have focused on optimizing the diffusion model itself. In the field of controlled generation guided by multimodal information, there is little related research on acceleration.

% 扩散模型的可控生成一直都是研究热点，其中controlnet作为一种即插即用的模块，被广泛推广使用。但是每个controlnet会增加推理过程大概一半的计算量，影响效率和算力成本。针对这一问题，我们受repvgg[]和controlnet[]工作启发，提出了一种模态重参数化方法RepControlNet。RepControlNet训练过程冻结原扩散模型，复制一份扩散模型的卷积层和线性层，用以学习输入的模态控制信息。推理过程将训练过程复制的卷积层和线性层和原扩散模型重参数化，不增加除扩散模型本身外的额外计算量的同时，实现条件控制生成。在sd1.5和sdxl上的实验结果证明了RepControlNet的有效性，相比于增加一半计算量的controlnet，条件控制生成能力也几乎没有损失。

Controllable generation of diffusion models has always been a research hotspot, with ControlNet \cite{zhang2023adding} being widely used as a plug-and-play module. However, each ControlNet increases the computational load of the inference process by about half, affecting efficiency and computational cost. To address this issue, inspired by RepVGG \cite{ding2021repvgg} and ControlNet \cite{zhang2023adding}, we propose a modal reparameterization method, RepControlNet. During the training process, RepControlNet freezes the original diffusion model and duplicates its convolutional and linear layers to learn the input modal control information. During inference, the duplicated convolutional and linear layers are reparameterized with the original diffusion model, achieving conditional controlled generation without adding extra computational load beyond the diffusion model itself. Experimental results on Stable Diffusion 1.5 (SD1.5) \cite{Rombach_2022_CVPR}  and Stable Diffusion XL (SDXL) \cite{podell2023sdxl} demonstrate the effectiveness of RepControlNet, showing that it maintains conditional control generation capabilities with almost no loss compared to ControlNet, which increases computational load by half.

% 1.针对扩散模型可控生成领域，提出了一种模态重参数化方法RepControlNet，能够在不增加计算量的前提下，实现条件控制生成。
% 2.在stable diffusion1.5和stable diffusion-XL基座上，验证了RepControlNet方法对于canny、depth、skeleton等模态的模态重参数化条件控制生成能力。

To sum up, our contributions are as following:

\begin{itemize}
\item In the field of controllable generation of diffusion models, we propose a modal reparameterization method, RepControlNet, realizing the conditional controllable generation without increasing the amount of computation.
\item Based on the SD1.5 and SDXL, we validate the modal reparameterization conditional control generation capabilities of the RepControlNet method for modalities such as canny, depth, semantic map and skeleton.
\end{itemize}

\section{Related Work}

\subsection{Controllable Diffusion Models}
% StableDiffusion, ControlNet, T2I-Adapter, IP-Adapter, InstantID
% StableDiffusion[]是图像生成领域的基础模型，提供了高质量的输出。然而，其缺乏可控性，限制了其在需要特定输出的场景中的应用。为了解决这一问题，ControlNet[]引入了即插即用的模块，通过增加额外的控制信号来指导生成过程。尽管ControlNet在可控生成方面取得了显著效果，但它显著增加了计算负担，使其效率较低。T2I-Adapter[]进一步扩展了这一思路，引入了文本到图像的可控性，允许基于文本输入进行更精确的生成。然而，T2I-Adapter也需要额外的参数，这在资源受限的环境中可能是一个限制。
% 与此同时，IP-Adapter[]专注于推理加速和模型压缩，使用适配器模块来调制输入信息。尽管IP-Adapter提高了速度和效率，但它仍然需要额外的参数。Instantid[]则提供了实例级别的可控性，使得在一个类别内生成特定实例成为可能。然而，Instantid也面临着计算复杂性增加的问题。
% 综上所述，尽管这些方法在可控生成方面取得了显著进展，但它们通常在计算负担和效率方面存在权衡。我们提出的RepControlNet旨在解决这些问题，在不增加计算复杂性的前提下实现可控生成。
Stable Diffusion \cite{Rombach_2022_CVPR} is a foundational model in the field of image generation, providing high-quality outputs. However, the lack of controllability limits its application in scenarios requiring specific outputs. To address this issue, ControlNet \cite{zhang2023adding} introduces plug-and-play modules that guide the generation process by adding additional control signals. Although ControlNet has achieved significant results in controllable generation, it significantly increases the computational burden, making it less efficient. T2I-Adapter \cite{mou2024t2i} further extends this idea by introducing text-to-image controllability, allowing for more precise generation based on text input. However, T2I-Adapter also requires additional parameters, which can be a limitation in resource-constrained environments. Meanwhile, IP-Adapter \cite{ye2023ip} focuses on inference acceleration and model compression, using adapter modules to modulate input information. Although IP-Adapter improves speed and efficiency, it still requires additional parameters. InstantID \cite{wang2024instantid} provides instance-level controllability, making it possible to generate specific instances within a category. However, InstantID also faces the issue of increased computational complexity.

Although these methods have made significant progress in controllable generation, they often involve trade-offs in terms of computational burden and efficiency. Our proposed RepControlNet aims to address these issues by achieving controllable generation without increasing computational complexity.

\subsection{Accelerating Diffusion Models}
% snapfution,步数蒸馏,repvgg
% 在扩散模型的加速方面，SnapFusion[]专注于模型的小型化，通过减少模型参数量和部署蒸馏技术来实现推理速度的提升。步数蒸馏通过压缩推理步骤的数量来加速推理过程。步数蒸馏的核心思想是通过训练一个教师模型来生成中间步骤的目标，然后用这些目标来训练一个学生模型，从而减少推理所需的步骤数量。例如，Consistency Models[]通过在训练过程中引入一致性约束，减少了推理步骤，同时保持了生成质量。Progressive Distillation[]采用渐进式蒸馏方法，通过逐步优化推理步骤，实现了高效的推理加速。然而，这些方法通常会牺牲一定的生成质量。

In the aspect of accelerating diffusion models, SnapFusion \cite{li2024snapfusion} focuses on model compression by reducing the amount of model parameters and deploying distillation techniques to enhance inference speed. Step distillation accelerates the inference process by compressing the number of inference steps. The core idea of step distillation is to train a teacher model to generate intermediate step targets, and then use these targets to train a student model, thereby reducing the number of steps required for inference. For example, Consistency Models \cite{song2023consistency} introduce consistency constraints during training to reduce inference steps while maintaining generation quality. Progressive Distillation \cite{salimans2022progressive} adopts a progressive distillation approach, optimizing inference steps incrementally to achieve efficient inference acceleration. However, these methods often come at the cost of sacrificing some generation quality.

% 重参数化技术是一种在训练和推理阶段优化模型架构的方法。训练阶段使用复杂的模型结构以提高模型的表达能力，而在推理阶段将这些复杂结构重参数化为更简单的形式，从而减少计算复杂度。RepVGG[]和RepViT[]是重参数化技术的典型应用。RepVGG通过将训练时的复杂卷积层重参数化为推理时的简单卷积层，从而减少计算复杂度。RepViT则将这一技术应用于Vision Transformer，在保持模型性能的同时显著提高了推理速度。然而，这些方法并未专门解决可控生成的问题。

Reparameterization techniques optimize model architecture during both training and inference stages. Complex model structures are used during training to enhance the model's expressive power, while these complex structures are reparameterized into simpler forms during inference to reduce computational complexity. RepVGG \cite{ding2021repvgg} and RepViT \cite{Wang_2024_CVPR} are typical applications of reparameterization techniques. RepVGG reparameterizes complex convolutional layers used during training into simple convolutional layers for inference, thereby reducing computational complexity. RepViT applies this technique to Vision Transformers, significantly improving inference speed while maintaining model performance. However, these methods do not specifically address the issue of controllable generation.

% 尽管这些方法在加速扩散模型方面做出了重要贡献，但它们通常仅关注模型本身，并未考虑可控生成所带来的额外复杂性。受RepVGG和ControlNet工作的启发,针对ControlNet显著增加计算开销的问题，我们提出了一种模态重参数化方法RepControlNet。通过在训练过程中冻结原扩散模型，并复制其卷积层和线性层以学习输入的模态控制信息，RepControlNet在推理过程中将这些复制的层与原扩散模型进行重参数化，从而在不增加计算量的前提下实现条件控制生成。

Although these methods have made significant contributions to accelerating diffusion models, they typically focus only on the model itself and do not consider the additional complexity introduced by controllable generation. Inspired by RepVGG and ControlNet, we propose a modal reparameterization method, RepControlNet, to address the issue of increased computational overhead in ControlNet. During the training process, RepControlNet freezes the original diffusion model and duplicates its convolutional and linear layers to learn the input modal control information. During inference, the duplicated convolutional and linear layers are reparameterized with the original diffusion model, achieving conditional controllable generation without adding extra computational load beyond the diffusion model itself.

\begin{figure}[t]
\begin{center}
\includegraphics[width=0.47\textwidth]{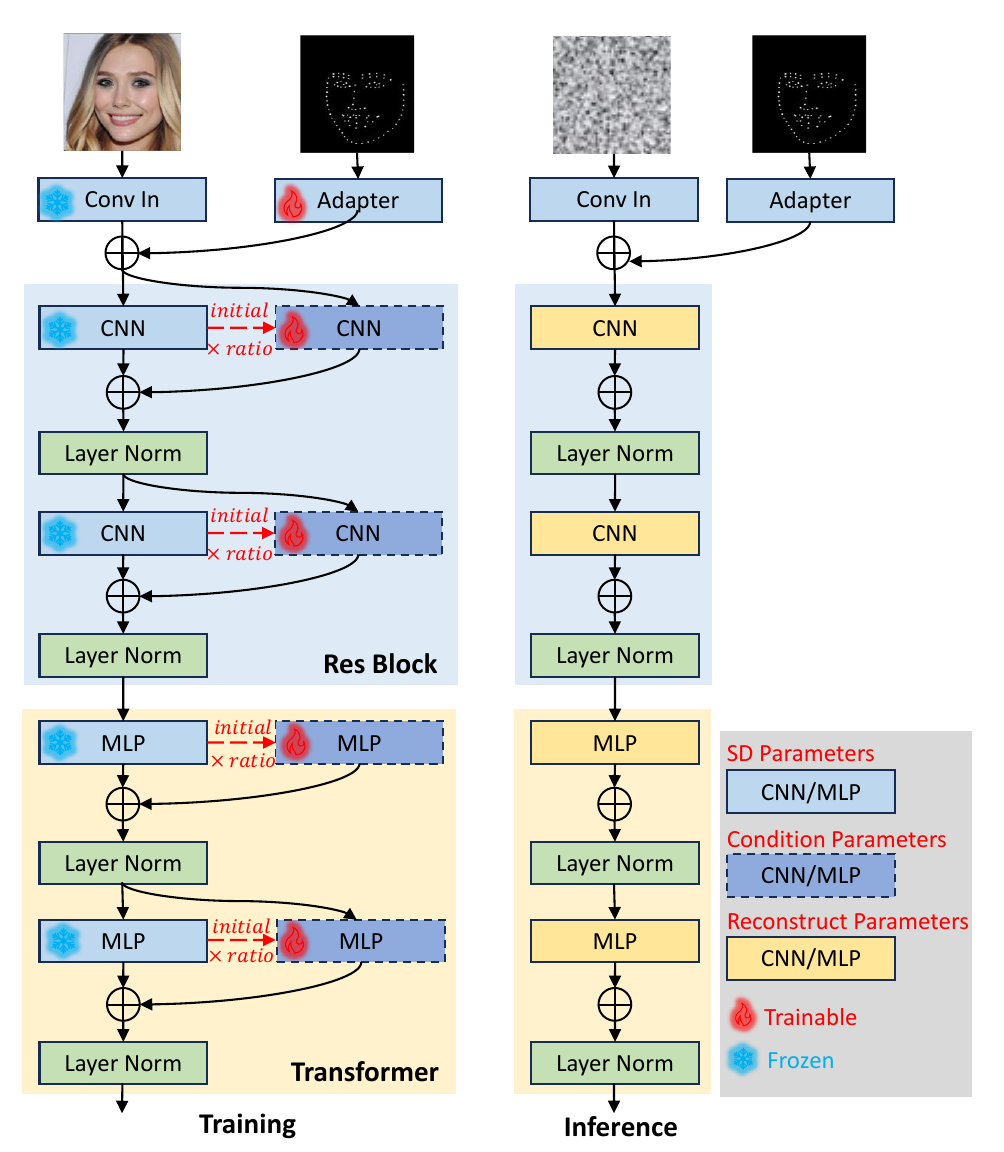}
\end{center}
\caption{Flowchart of training and reparameter process of RepControNet.}
\label{fig:flowchart}
\end{figure}

\section{Methods}
\subsection{Preliminaries}
% 介绍ControlNet/Adapter技术
% 介绍Reparamterization
% ControlNet将额外的条件注入到预训练的Diffusion网络的块中。通过锁定（冻结）原始块的参数 $\Theta$，并同时克隆该块为一个可训练的副本 $\Theta_c$，可训练的副本接受一个外部条件向量 $c$ 作为输入，并通过零卷积层与锁定的模型连接，其计算如下：
% $$
% y_c = F(x; \Theta) + Z(F(x + Z(c; \Theta_{z1}); \Theta_c); \Theta_{z2})
% $$
% 在Stable Diffusion中，ControlNet结构应用于U-Net backbone的每个编码器级别，使用ControlNet创建Stable Diffusion的12个编码块和1个中间块的可训练副本。由于锁定的副本参数被冻结，在微调过程中不需要在原始锁定的编码器中进行梯度计算。这种方法加快了训练速度并节省了GPU内存。
% 然而，ControlNet创建的可训练副本增加了计算开销和模型参数量，从而影响了推理过程的效率。
% 目前，许多研究已经致力于优化模型架构，以提高其性能和效率。其中，**重参数化技术**是一种在训练和推理阶段优化模型架构的有效方法。训练阶段使用复杂的模型结构以提高模型的表达能力，而在推理阶段将这些复杂结构重参数化为更简单的形式，从而减少计算复杂度。**RepVGG**是重参数化技术的典型应用。**RepVGG**通过在训练阶段使用多分支拓扑结构，并在推理阶段将其重参数化为简单的平面架构，从而显著减少计算复杂度，在速度和准确性之间实现了良好的权衡。
% 因此，受**RepVGG**的启发，我们提出了一种模态重参数化方法**RepControlNet**，在不增加除扩散模型本身外的额外计算量的同时，实现条件控制生成。
In ControlNet, the control element $c$ is injected into Diffusion model by the trainable copy $\Theta_c$ of original blocks $\Theta$ which is frozen in training process, and zero convolutions.
\begin{equation}
\label{controlnet}
\begin{aligned}
y_c = F(x; \Theta) + Z(F(x + Z(c; \Theta_{z1}); \Theta_c); \Theta_{z2})
\end{aligned}
\end{equation}
In Stable Diffusion, ControlNet is applied in the 12 encoder blocks and 1 middle block of U-Net. 
The gradients are not calculated in the original encoder blocks frozen in the training process, which speeds up the training phase and saves GPU memory. 
However, it reduces the computational efficiency in the inference by increasing the number of parameter and the computational overhead.
Currently, many methods focus on the optimization of model architecture, aiming to increase the model performance and inference efficiency.  
Reparameterization is one of the most efficient methods among them.
The complicated model is used to improve the capacity of model in the training process, while the reparameterization leading to less computation is applied to simplify the model in the inference process. 
RepVGG \cite{ding2021repvgg} is the typical method of reparameterization. RepVGG uses the multi-branch topology structure in the training process and is reparametered as simple plane structure in the inference process, which makes a balance between inference speed and accuracy.
Inspired by RepVGG, we propose a modal-reparameter method RepControlNet, which realizes the equivalent effect as ControlNet without increasing additional computational complexity of diffusion model.

% \begin{figure}[ht]
% \begin{center}
% \includegraphics[width=0.48\textwidth]{figs/RepControlNet.pdf}
% % \includegraphics[width=0.48\textwidth]{figs/flowchart.jpg}
% \end{center}
% \caption{Flowchart of training and reparameter process of RepControNet.}
% \label{fig:flowchart}
% \end{figure}

\subsection{Structure}
% 与ControlNet类似，**RepControlNet**在训练过程中锁定（冻结）原始块的参数，并同时克隆该块为一个可训练的副本，用以学习输入的模态控制信息。

% 然而，与ControlNet不同的是，为了在推理过程中利用重参数化技术，我们仅对UNet中的卷积层和线性层创建副本。

% 具体来说，**RepControlNet**在训练阶段首先锁定预训练扩散模型中所有卷积层和线性层的参数，以确保这些参数在训练过程中保持不变。然后，克隆这些冻结的卷积层和线性层，创建一个可训练的副本。如FIG~\ref{fig:flowchart}所示，在**RepControlNet**的每个卷积层和线性层中，输入的特征张量同时输入给原始层和副本层，输出为两者的叠加。在UNet的第一层中，我们通过一个Adapter将输入条件进行编码，得到与UNet相同维度的输入张量，并与原始输入张量叠加，从而注入条件信息。

% 这样，原始的单路径结构变成了多分支结构，训练过程对可训练的副本进行优化，能够保证与ControlNet类似的高性能和学习能力。

% 为了重用大规模预训练模型的能力，并且使得训练初期多分支模型与原始模型不会产生太大的偏差，我们在初始化可训练的副本的参数时，将其置为原始块的参数乘以一个常数，即

% $$
% \text{Init}(\Theta_m) = w\Theta
% $$

% 这里我们经验性地将常数 $w$ 设置为0.1。
Similar to ControlNet, the original blocks in diffusion model are frozen in RepControlNet, while a trainable copy is applied to learn conditions. Unlike ControlNet, the trainable copy is only applied to convolution and linear layers in RepControlNet for the convenient reparameterization in inference process.
Specifically, in RepControlNet, all convolution and linear layers are frozen in pretrained diffusion model in training process, then the copy of these layers is trained as Fig.~\ref{fig:flowchart}.
In RepControlNet, the feature embedding is simultaneously input into original layers and copy layers in each convolution and linear layers, then adding the two outputs together as the output feature. 
An Adapter layer is applied to encode the input condition in the first layer of UNet. 
The condition information is injected by adding Adapter output tensor and original input tensor (the two tensors have the same dimension).

In this way, we transfer the single branch model to multi-branch model which achieves the equivalent effect as ControlNet. 
For reusing the strong capacity of pretrained diffusion model and reducing the gap between multi-branch model and original model as much as possible
in the early training process, the weights $\Theta_m$ in multi-branch model is initialized as a multiple of original weights $\Theta$ and a ratio $w$
\begin{equation}
\label{reparameterization}
\begin{aligned}
\Theta_m = w\Theta ,
\end{aligned}
\end{equation}
empirically, the ratio $w$ is set to be $0.1$ in our experiments.

\subsection{Reparameter}

% **RepControlNet**的独特之处在于其采用了重参数化技术，以在推理过程中实现高效的条件控制生成。在推理阶段，**RepControlNet**通过重参数化技术将训练时的复杂结构简化为高效的推理结构。具体来说，卷积层和线性层的副本与原始扩散模型的权重进行重参数化，通过将训练时学习到的权重与原始权重进行线性组合，实现条件控制生成，而不增加额外的计算量。

% 具体而言，对于每个卷积层和线性层，RepVGG中已经证明了通过组合不同分支的权重，能够将多分支模型等效转化为单路径模型，重参数化过程中的权重组合公式如下：
% $$
% \Theta' = \alpha \Theta + \beta \Theta_m
% $$

% 其中，$\Theta$是原始扩散模型的权重，$\Theta_m$是模态网络的权重，$\alpha$和$\beta$是控制权重比例的超参数。

% 通过上述方法，我们将推理时的**RepControlNet**转化为与原始扩散模型相同参数量的模型，从而在不增加额外计算量的同时，实现了高效的条件控制生成。

The uniqueness of RepControlNet lies in its use of reparameterization technology to achieve efficient conditional controllable generation during the inference process.
In the inference process, RepControlNet simplifies the complex structure during training into an efficient inference structure through reparameterization. 
Specifically, the copy of convolution and linear layers are reparameterized with the weights of the original diffusion model. 
By linearly combining the weights learned during training process with the original weights, conditional controllable generation is achieved without increasing additional computational complexity

Specifically, for each convolution layer and linear layer, RepVGG has demonstrated that by combining the weights of different branches, a multi-branch model can be equivalently transformed into a single-branch model. The weight combination formula in the reparameterization process is 
\begin{equation}
\label{reparameterization}
\begin{aligned}
\Theta' = \alpha \Theta + \beta \Theta_m, 
\end{aligned}
\end{equation}
where $\Theta$ is the weight of original diffusion model, $\Theta_m$ is the weight of the copy (modal) model, $\alpha$ and $\beta$ are the ratio hyper-parameters.

Through the above method, we have transformed the RepControlNet into a model with the same number of parameters as the original diffusion model during inference process, thereby achieving efficient conditional control generation without increasing additional computational complexity

\subsection{Identity-Preserving Generation}
Customized character generation is a trending topic in the field of image synthesis. We can also train the RepControlNet to produce realistic images of a given identity and posture. We combine RepControlNet with our Self-developed portrait fidelity generation method (not open source yet). During training, the control conditions for the images are based on human skeletens, while we incorporate identity information using cross attention. We utilize Clip Vision \cite{DBLP:journals/corr/abs-2103-00020}) and ArcFace~\cite{deng2018arcface} encoders to extract fine-grained face features. Then resample these features into face embedding $F_{id}$ with the dimension of 4$\times$768. $F_{id}$ are injected into RepControlNet by cross attention, the specific operation is as show in Eq.~\ref{crossattn_id_func},
\begin{equation}
\label{crossattn_id_func}
\begin{aligned}
CA =\operatorname{CrossAttn}\left(Q, F_{text} \right) + \operatorname{CrossAttn}\left(Q, F_{id}\right),\\
\end{aligned}
\end{equation}
where $CA$ is the cross attention feature in diffusion model. $Q$ is the attention query of RepControlNet, with skeleton as the input conditional image, $F_{text}$ is the text embedding.

During inference, we reparameterize the RepControlNet, whereas the introduction of identity information cannot be reparameterized. However, consistent with $F_{text}$, $F_{id}$ needs to be extracted only once, independent of time steps. Moreover, due to the limited number of tokens in  $F_{id}$, the additional computational overhead for inference is negligible.
%\MGFF alg
% \begin{algorithm}[tb]
% \caption{Mask Guided Face Fusion}
% \label{alg:algorithm}
% \textbf{Input}: $Q_{cross}$, \\
% Text Feature Embedding $F_{text}$, \\
% Face Region Mask $M$=$\{M_{id}^{1},{M_{id}^{2},...,{M_{id}^{N}\}$,\\
% ID Visual Feature Embedding $F_{id}$=$\{F_{id}^{1},F_{id}^{2},...,F_{id}^{N}\}$,  \\
% \textbf{Parameter}: $N$, the numer of ip\\
% \textbf{Output}: $A_{out}$, the attention map
% \begin{algorithmic}[1] %[1] enables line numbers
% \STATE \textbf $A_{text} \gets CrossAttn(Q_{cross}, F_{text})$\\

% \FOR{$i = 1$ to $N$ }
%     \STATE \textbf$A_{id}^{i} \gets CrossAttn(x, F_{id}^{i})\\
%     \STATE $A_{id}^{i} \gets A_{id}^{i} \odot M_{id}^{i}$\\
% \ENDFOR
% \STATE \textbf $A_{out} = A_{text} + {\textstyle \sum_{i=1}^{N}}A_{id}^{i} $
% \STATE \textbf{return}  $A_{out}$
% \end{algorithmic}
% \end{algorithm}

\begin{figure}[!t]
\centering
\begin{minipage}[c]{1\columnwidth}
    \makebox[0.32\columnwidth]{Canny edge}
    \makebox[0.32\columnwidth]{Original RGB}
    \makebox[0.32\columnwidth]{Generated}
\end{minipage}
\\
\includegraphics[width=0.32\columnwidth]{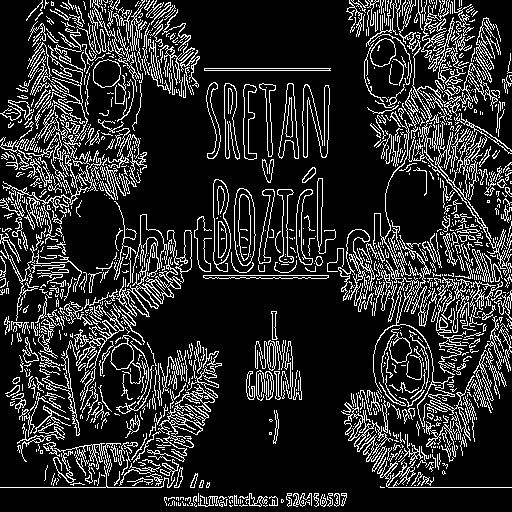}
\includegraphics[width=0.32\columnwidth]{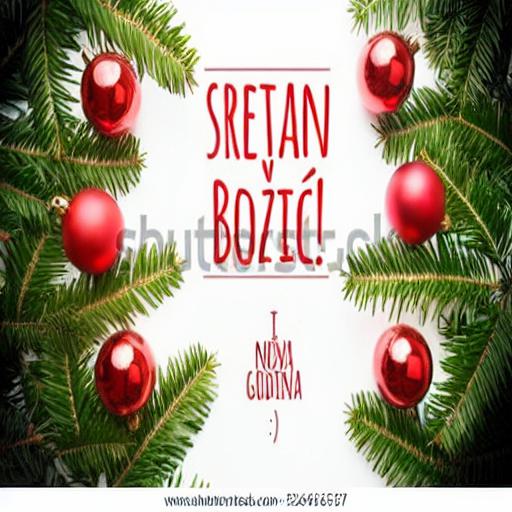}
\includegraphics[width=0.32\columnwidth]{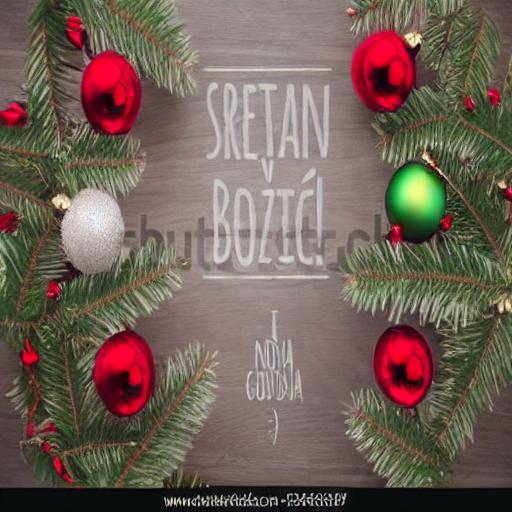}
\\
\includegraphics[width=0.32\columnwidth]{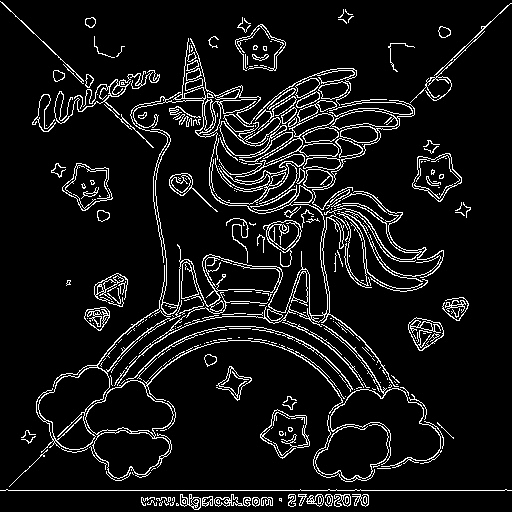}
\includegraphics[width=0.32\columnwidth]{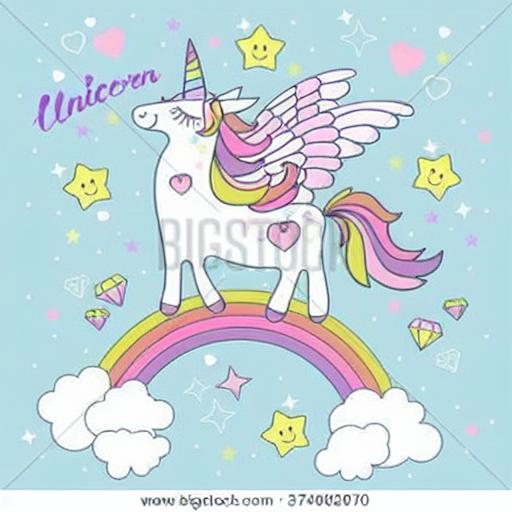}
\includegraphics[width=0.32\columnwidth]{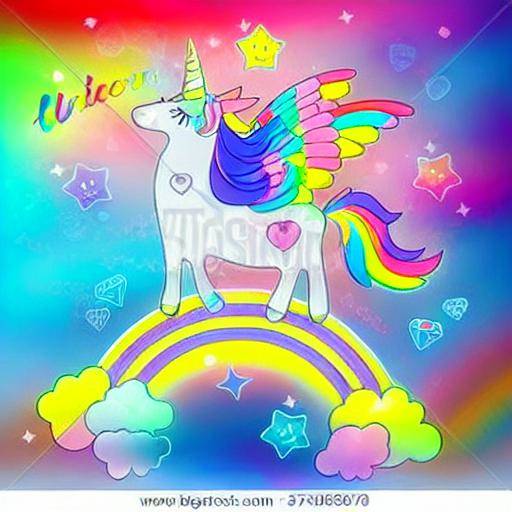}
\\
\includegraphics[width=0.32\columnwidth]{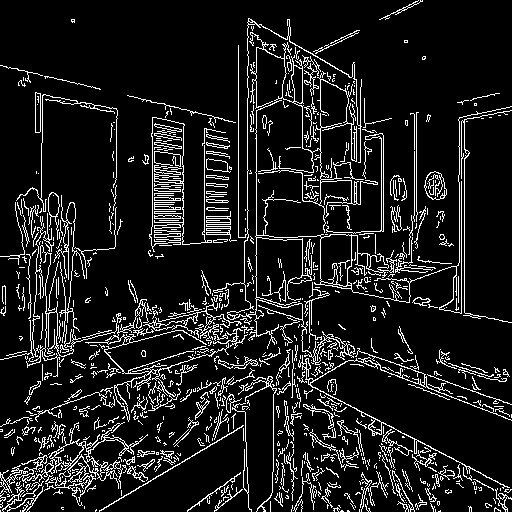}
\includegraphics[width=0.32\columnwidth]{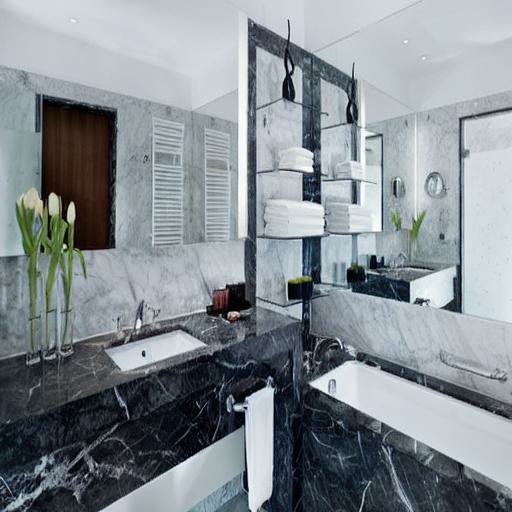}
\includegraphics[width=0.32\columnwidth]{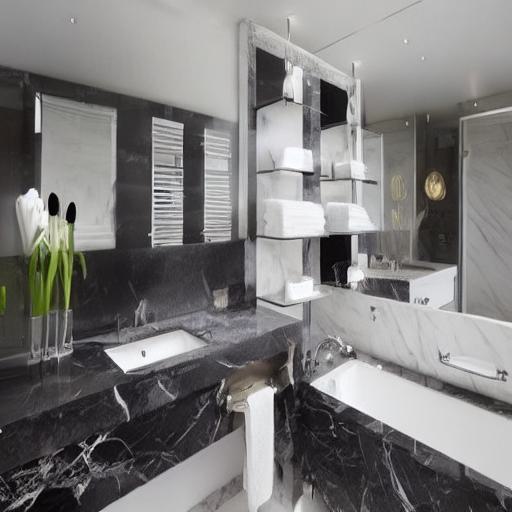}
\\
\caption{Image generation results with \textbf{Canny Edge} conditioning. Captions: Christmas greeting with wreath and red bauble on white background, scandinavian christmas text. A beautiful unicorn with a rainbow on a background of clouds unicorn cartoon character for pringting posters. The marble bathroom feature white and black marble walls and floors.}
\label{fig:canny}
\end{figure}

\begin{figure}[!t]
\centering
\begin{minipage}[c]{1\columnwidth}
    \makebox[0.32\columnwidth]{Depth map}
    \makebox[0.32\columnwidth]{Original RGB}
    \makebox[0.32\columnwidth]{Generated}
\end{minipage}
\\
\includegraphics[width=0.32\columnwidth]{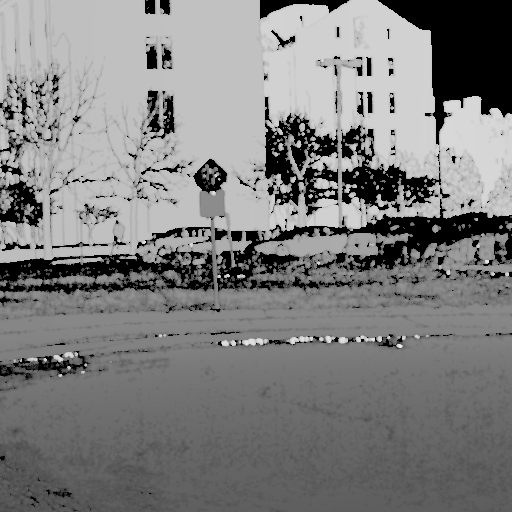}
\includegraphics[width=0.32\columnwidth]{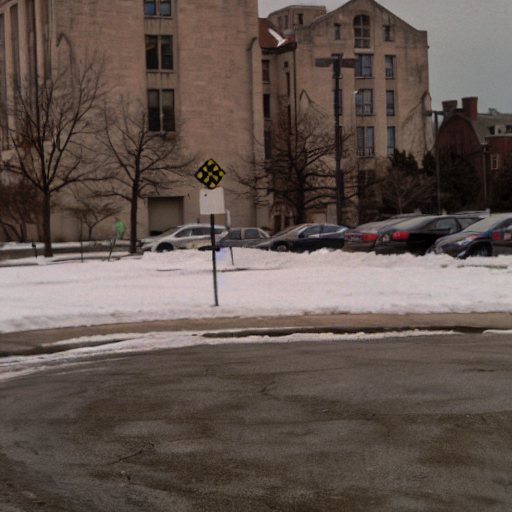}
\includegraphics[width=0.32\columnwidth]{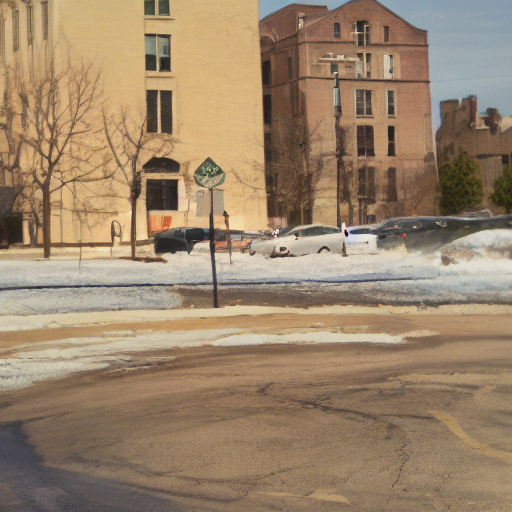}
\\
\includegraphics[width=0.32\columnwidth]{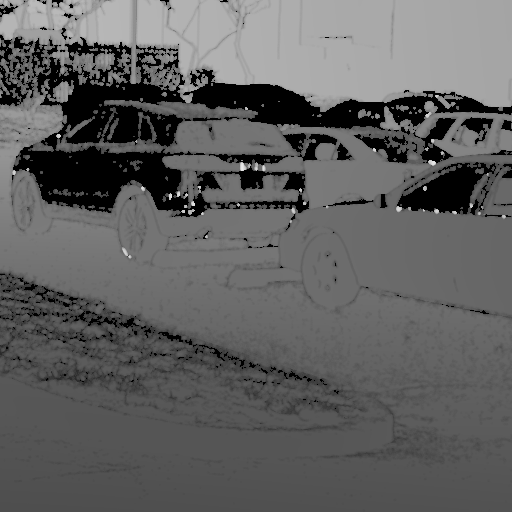}
\includegraphics[width=0.32\columnwidth]{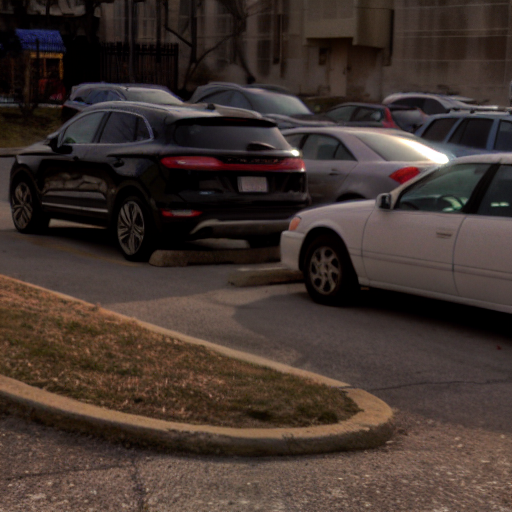}
\includegraphics[width=0.32\columnwidth]{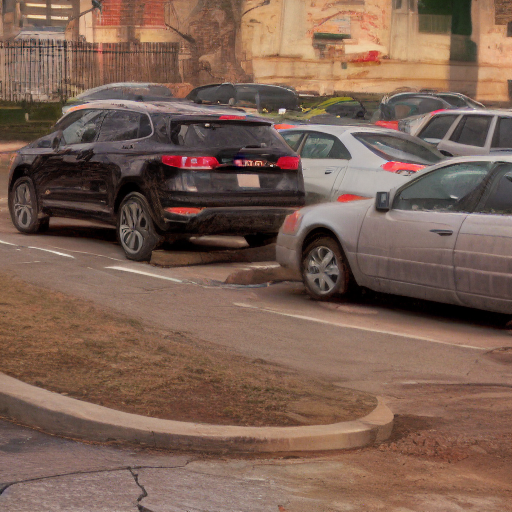}
\\
\includegraphics[width=0.32\columnwidth]{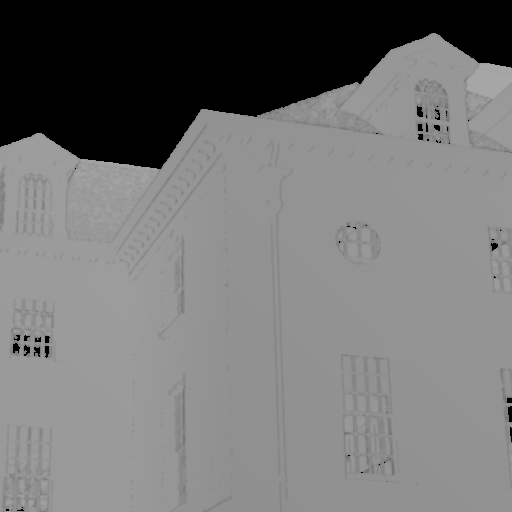}
\includegraphics[width=0.32\columnwidth]{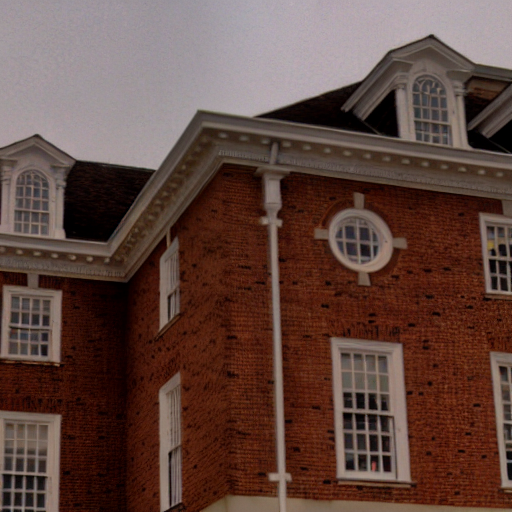}
\includegraphics[width=0.32\columnwidth]{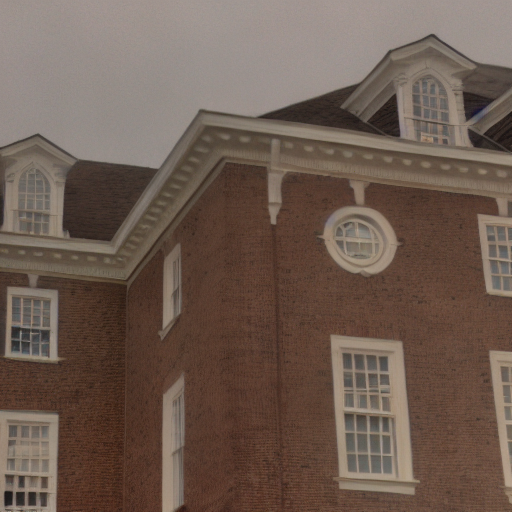}
\caption{Image generation results with \textbf{Depth Map} conditioning. Captions: A parking lot with a stop sign and a building. A car parked by a red brick wall. A large red brick building.}
\label{fig:depth}
\end{figure}

\begin{figure}[!t]
\centering
\begin{minipage}[c]{1\columnwidth}
    \makebox[0.32\columnwidth]{COCO-Stuff seg}
    \makebox[0.32\columnwidth]{Original RGB}
    \makebox[0.32\columnwidth]{Generated}
\end{minipage}
\\
\includegraphics[width=0.32\columnwidth]{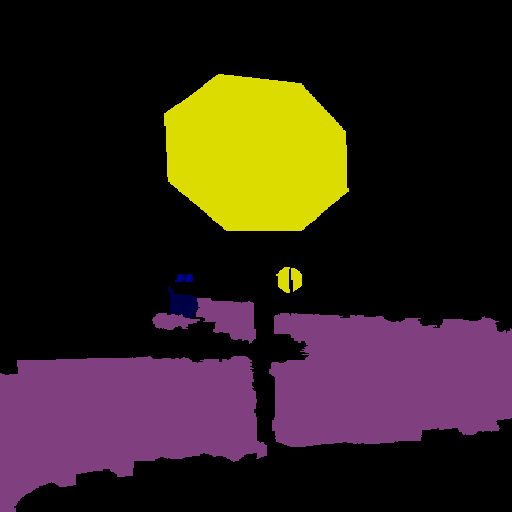}
\includegraphics[width=0.32\columnwidth]{figs/SS_COCO/reconstruction/000000000724.jpg}
\includegraphics[width=0.32\columnwidth]{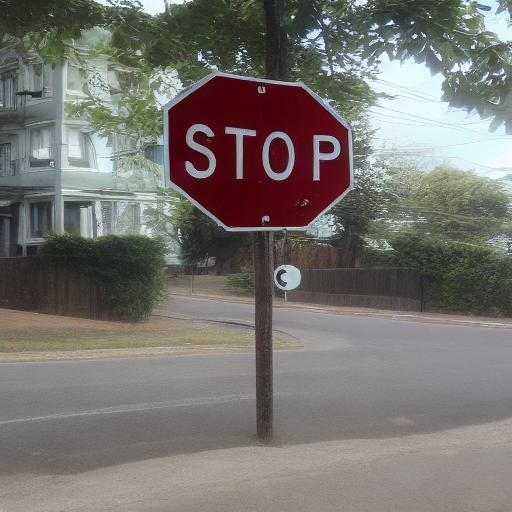}
\\
\includegraphics[width=0.32\columnwidth]{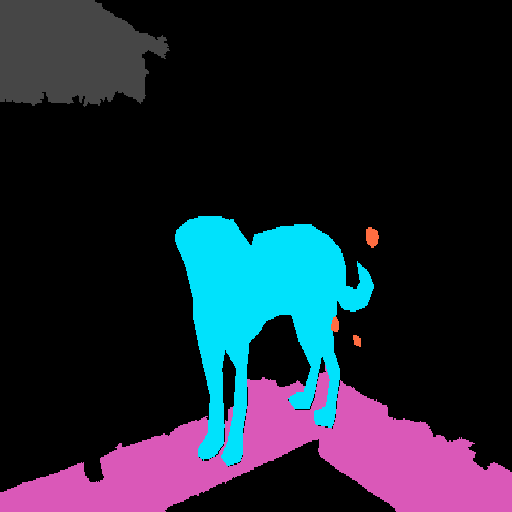}
\includegraphics[width=0.32\columnwidth]{figs/SS_COCO/reconstruction/000000029393.jpg}
\includegraphics[width=0.32\columnwidth]{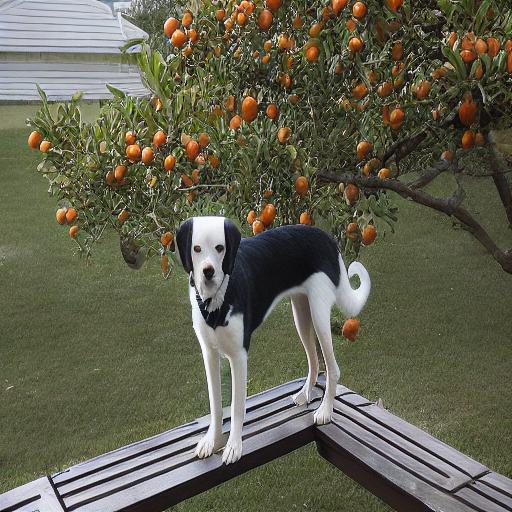}
\\
\includegraphics[width=0.32\columnwidth]{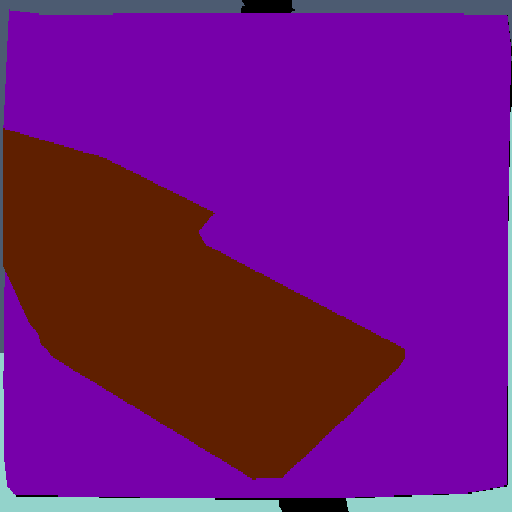}
\includegraphics[width=0.32\columnwidth]{figs/SS_COCO/reconstruction/000000134112.jpg}
\includegraphics[width=0.32\columnwidth]{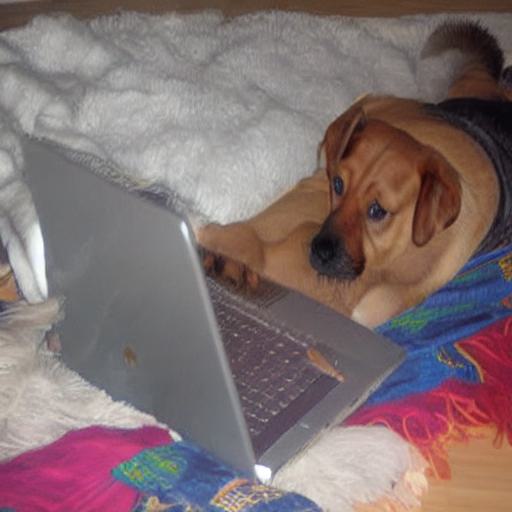}
\caption{Image generation results with \textbf{COCO segmentation} conditioning. Captions: An upside down stop sign by the road. A dog standing on a bench with an orange tree in back. A dog is lying next to a laptop computer.}
\label{fig:ss_coco}
\end{figure}

\begin{figure}[!t]
\centering
\begin{minipage}[c]{1\columnwidth}
    \makebox[0.24\columnwidth]{ADE seg}
    \makebox[0.24\columnwidth]{Original RGB}
    \makebox[0.24\columnwidth]{Generated}
    \makebox[0.24\columnwidth]{Generated}
\end{minipage}
\\
\begin{minipage}[c]{1\columnwidth}
    \makebox[0.24\columnwidth]{}
    \makebox[0.24\columnwidth]{}
    \makebox[0.24\columnwidth]{(SD1.5)}
    \makebox[0.24\columnwidth]{(SDXL)}
\end{minipage}
\\
\includegraphics[width=0.24\columnwidth]{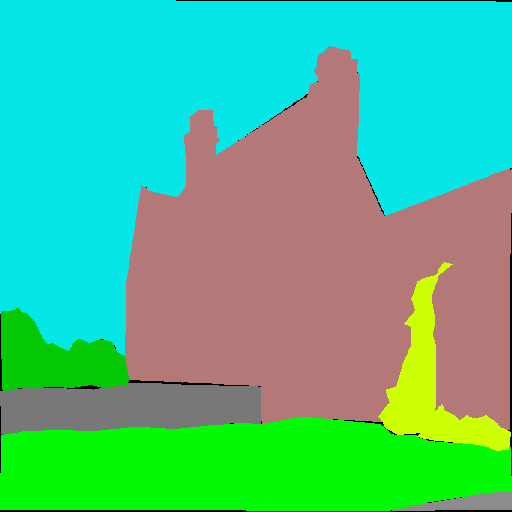}
\includegraphics[width=0.24\columnwidth]{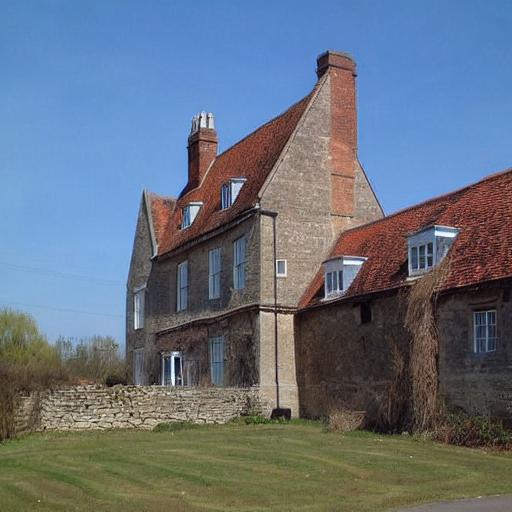}
\includegraphics[width=0.24\columnwidth]{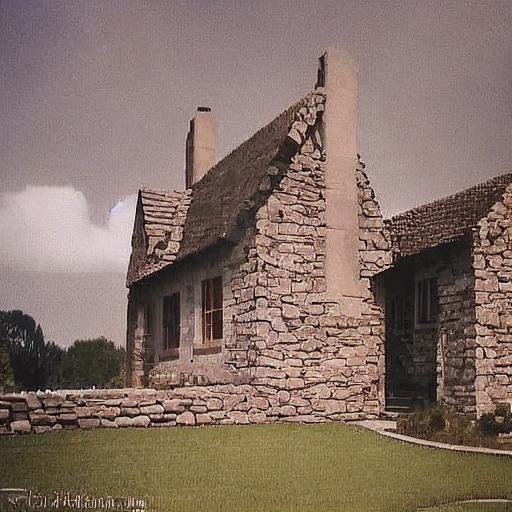}
\includegraphics[width=0.24\columnwidth]{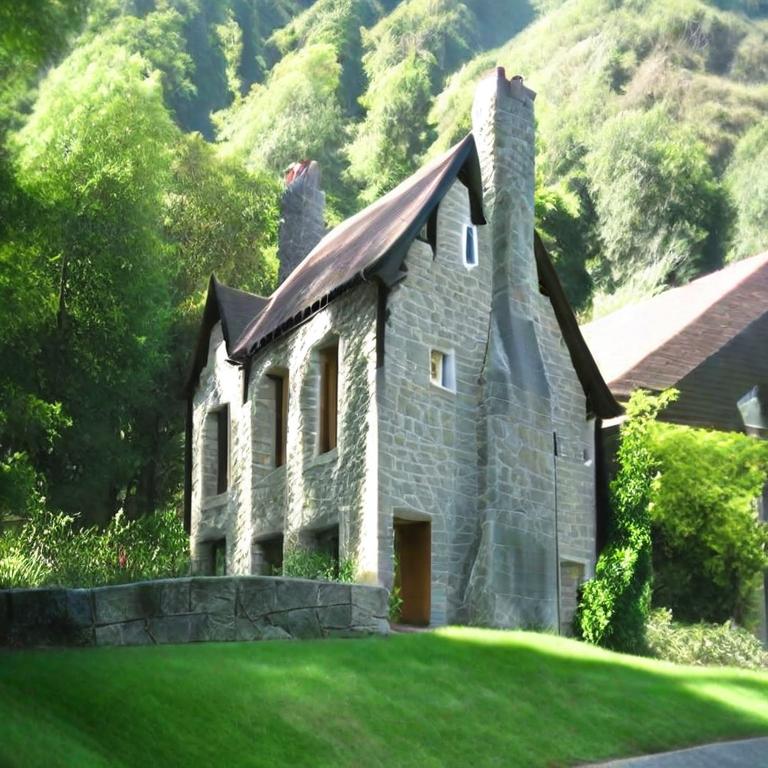}
\\
\includegraphics[width=0.24\columnwidth]{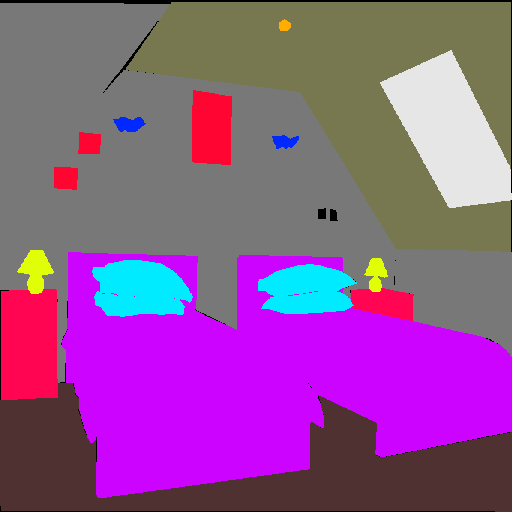}
\includegraphics[width=0.24\columnwidth]{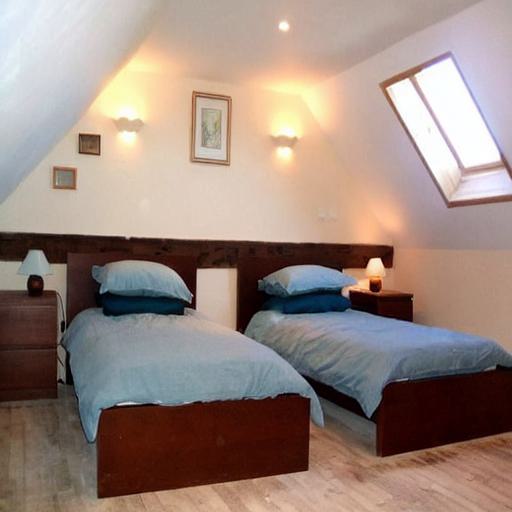}
\includegraphics[width=0.24\columnwidth]{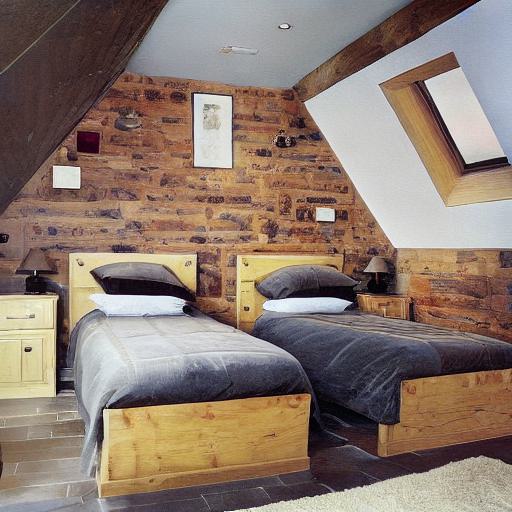}
\includegraphics[width=0.24\columnwidth]{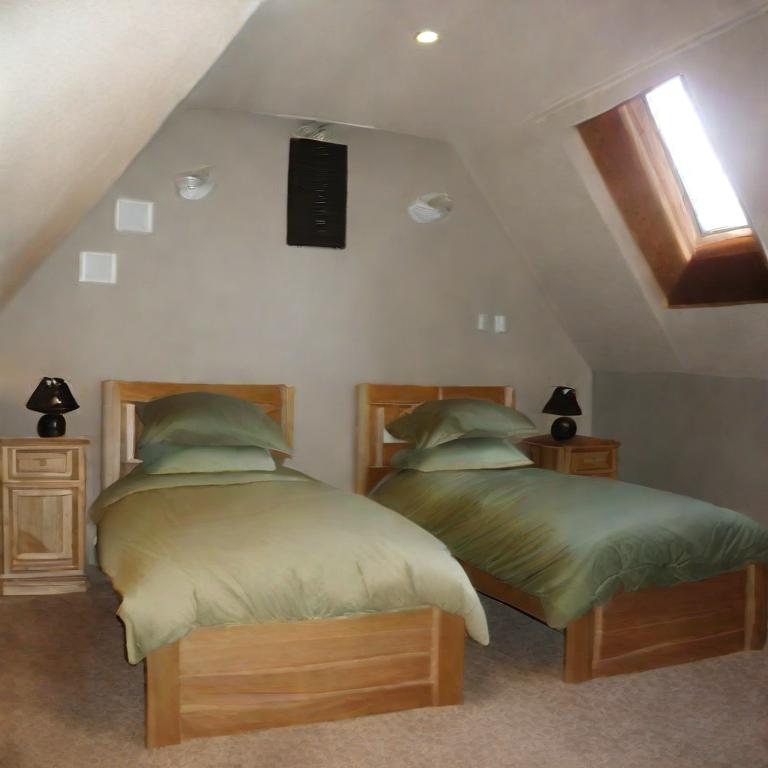}
\\
\includegraphics[width=0.24\columnwidth]{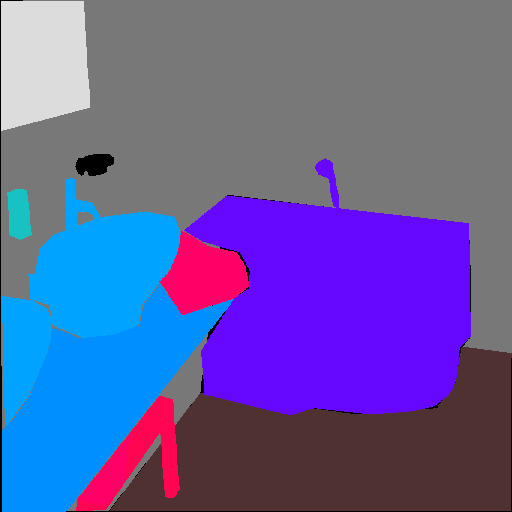}
\includegraphics[width=0.24\columnwidth]{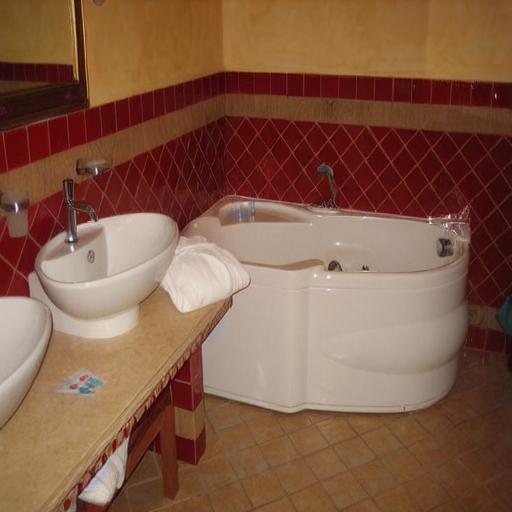}
\includegraphics[width=0.24\columnwidth]{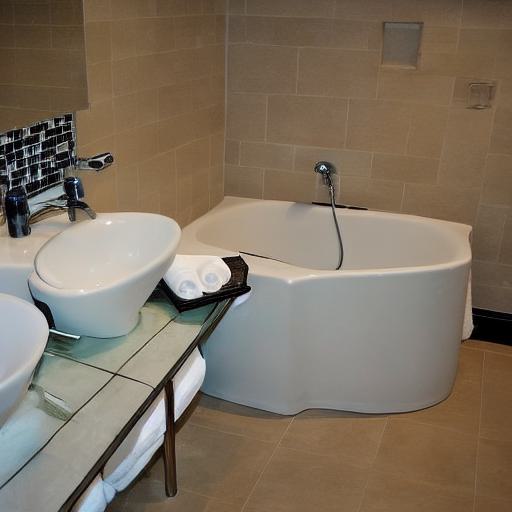}
\includegraphics[width=0.24\columnwidth]{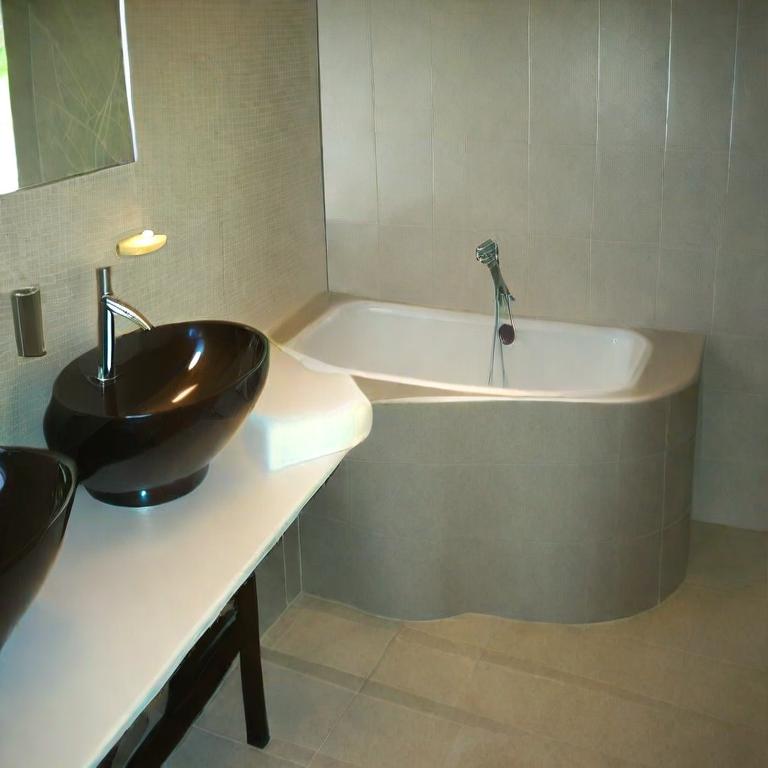}
\\
\includegraphics[width=0.24\columnwidth]{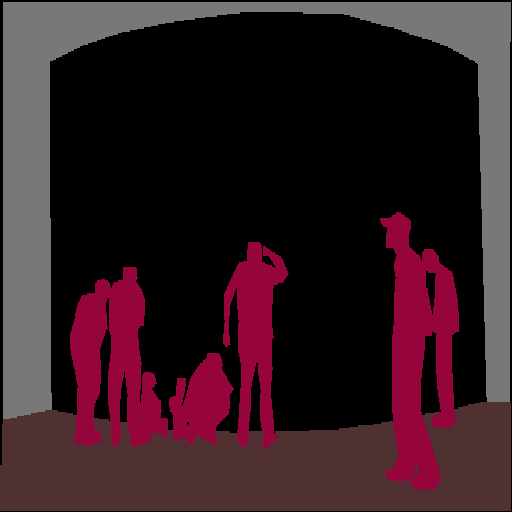}
\includegraphics[width=0.24\columnwidth]{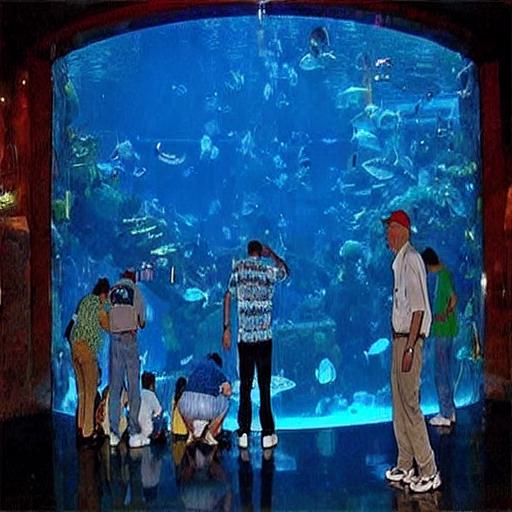}
\includegraphics[width=0.24\columnwidth]{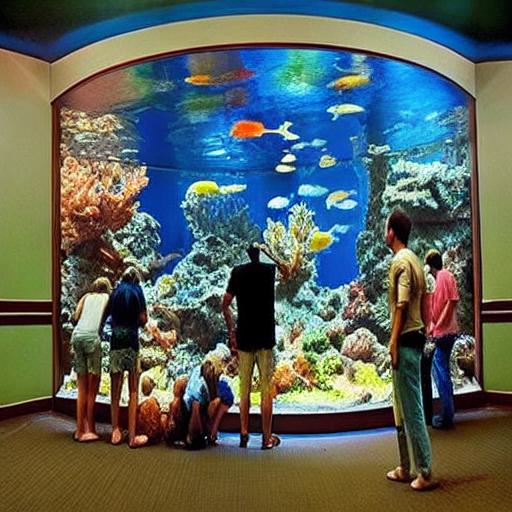}
\includegraphics[width=0.24\columnwidth]{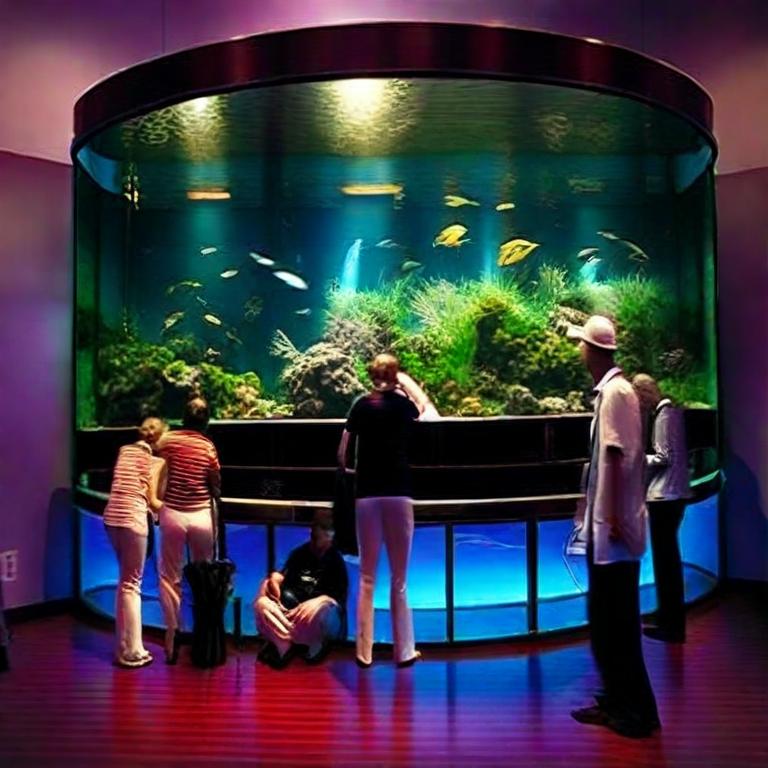}
\caption{Image generation results with \textbf{ADE segmentation} conditioning. Captions: A large stone house. Two beds in a bedroom. A bathroom with two sinks and a tub. People looking at an aquarium with fish in it.}
\label{fig:ss_ade}
\end{figure}

\begin{figure*}[!t]
\centering
\begin{minipage}[c]{0.16\textwidth}
    \makebox[1\textwidth]{Conditioning} \\
        \makebox[1\textwidth]{} \\
    \includegraphics[width=1\textwidth]{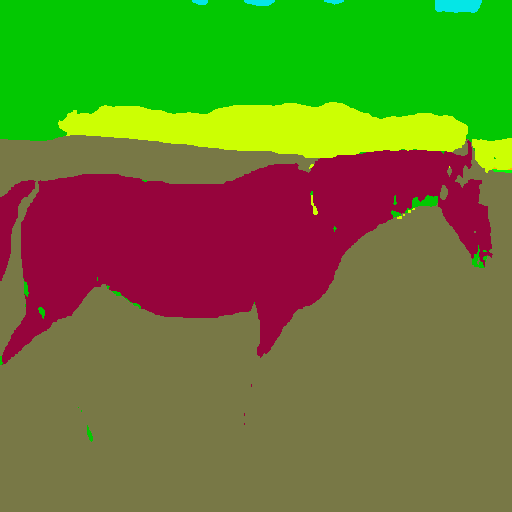}\\
    \includegraphics[width=1\textwidth]{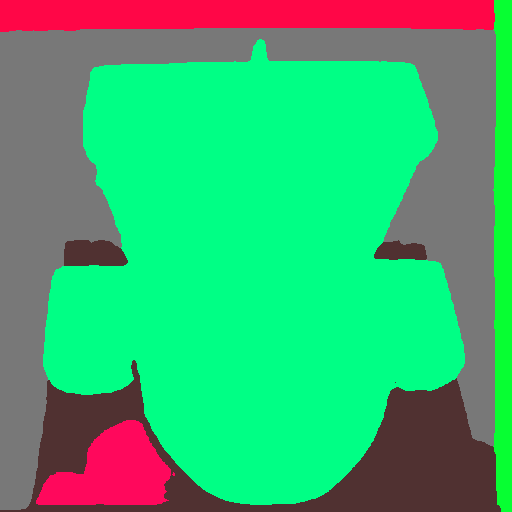}\\
    \includegraphics[width=1\textwidth]{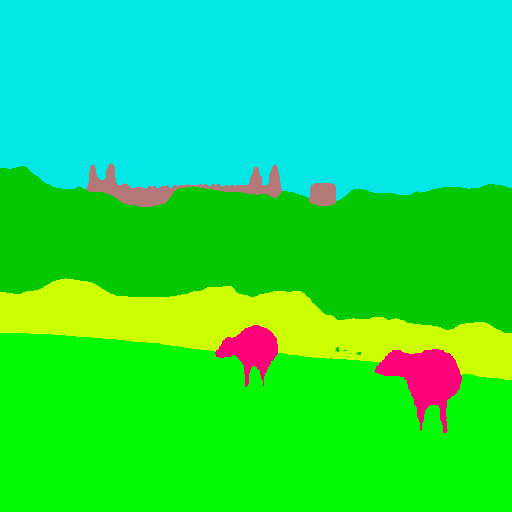}\\
    \includegraphics[width=1\textwidth]{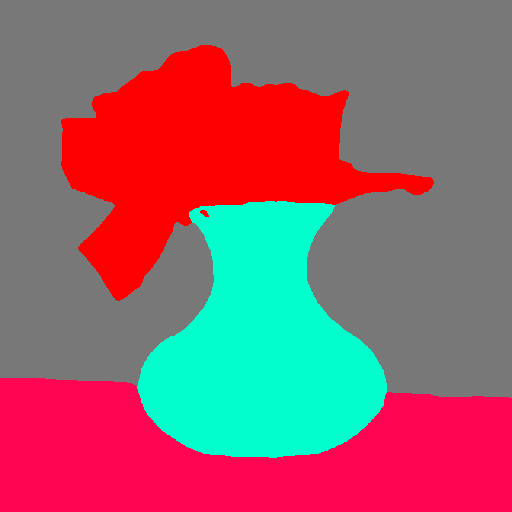}\\
    \includegraphics[width=1\textwidth]{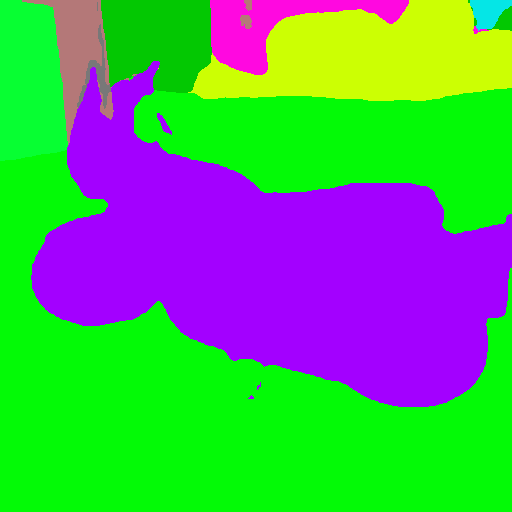}
\end{minipage}
\begin{minipage}[c]{0.16\textwidth}
    \makebox[1\textwidth]{Original RGB} \\
    \makebox[1\textwidth]{} \\
    \includegraphics[width=1\textwidth]{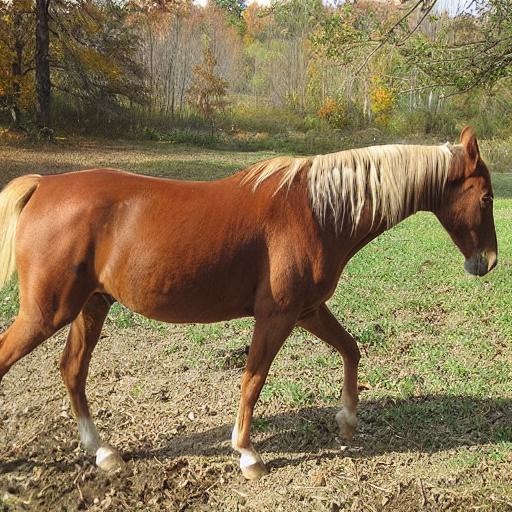}\\
    \includegraphics[width=1\textwidth]{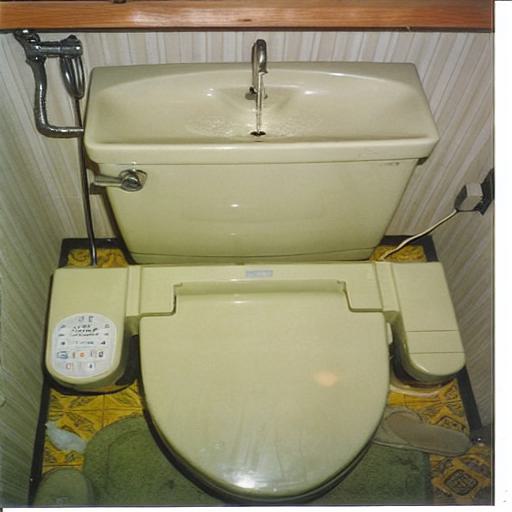}\\
    \includegraphics[width=1\textwidth]{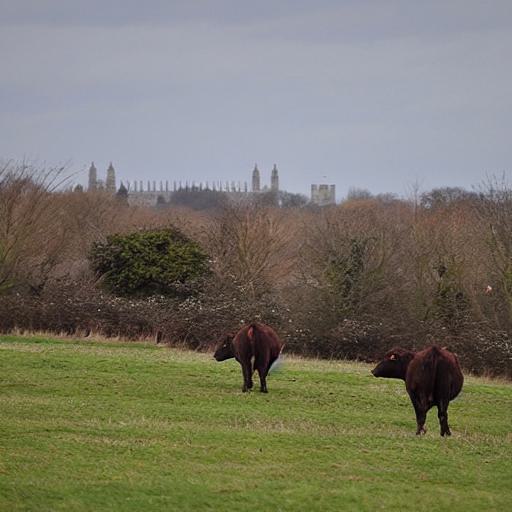}\\
    \includegraphics[width=1\textwidth]{figs/SS_COCO_10000/our_ade/reconstruction/19468.jpg}\\
    \includegraphics[width=1\textwidth]{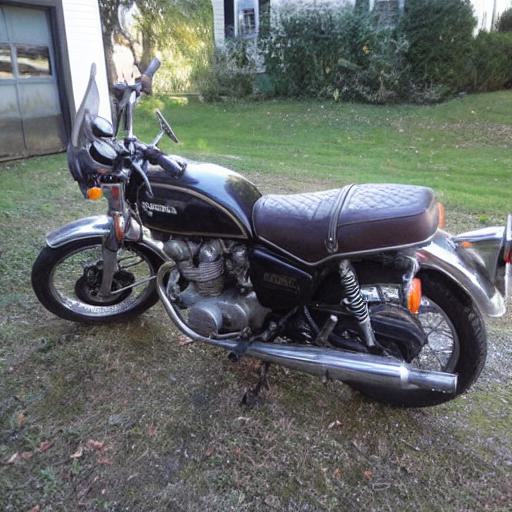}
\end{minipage}
\begin{minipage}[c]{0.16\textwidth}
    \makebox[1\textwidth]{ControlNet} \\
    \makebox[1\textwidth]{(SD1.5)} \\
    \includegraphics[width=1\textwidth]{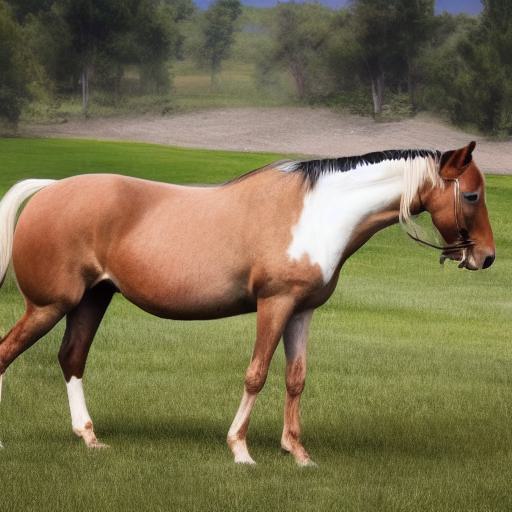}\\
    \includegraphics[width=1\textwidth]{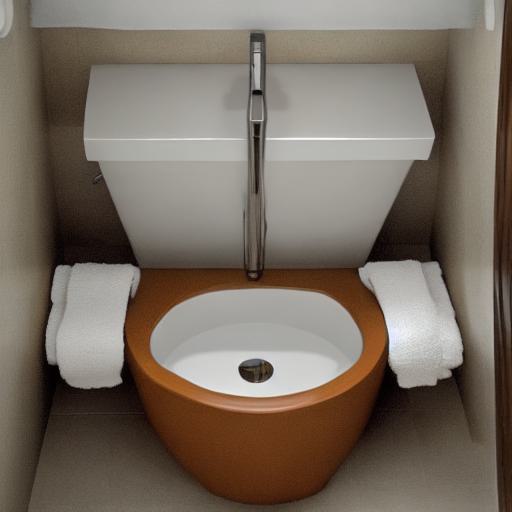}\\
    \includegraphics[width=1\textwidth]{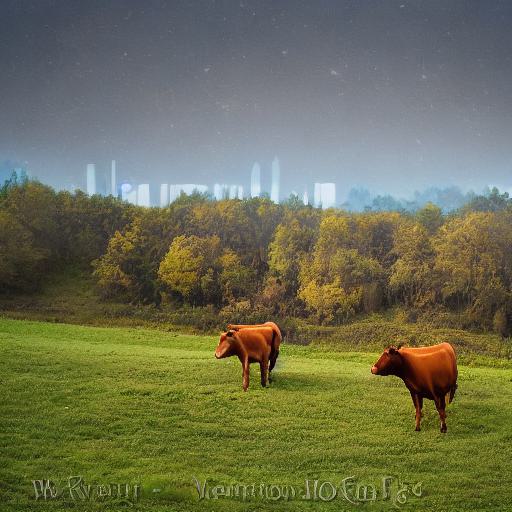}\\
    \includegraphics[width=1\textwidth]{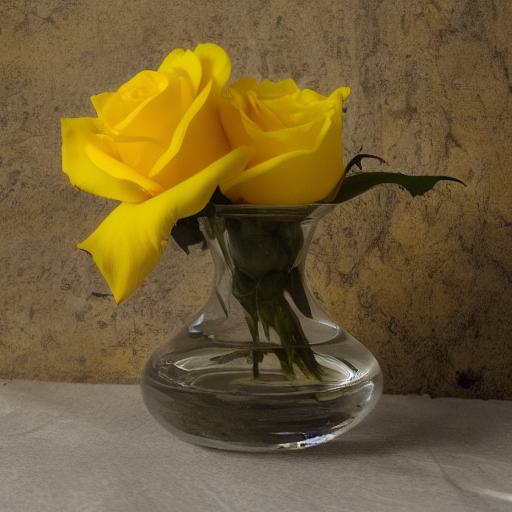}\\
    \includegraphics[width=1\textwidth]{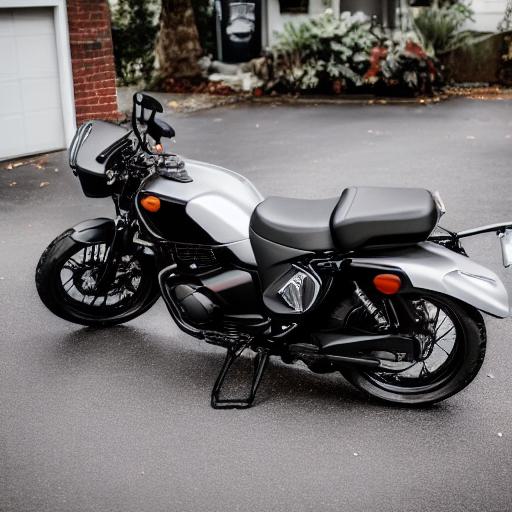}
\end{minipage}
\begin{minipage}[c]{0.16\textwidth}
    \makebox[1\textwidth]{RepControlNet+} \\
    \makebox[1\textwidth]{(SD1.5)} \\
    \includegraphics[width=1\textwidth]{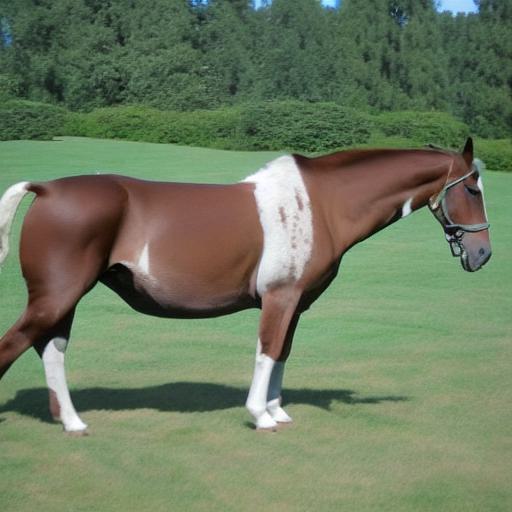}\\
    \includegraphics[width=1\textwidth]{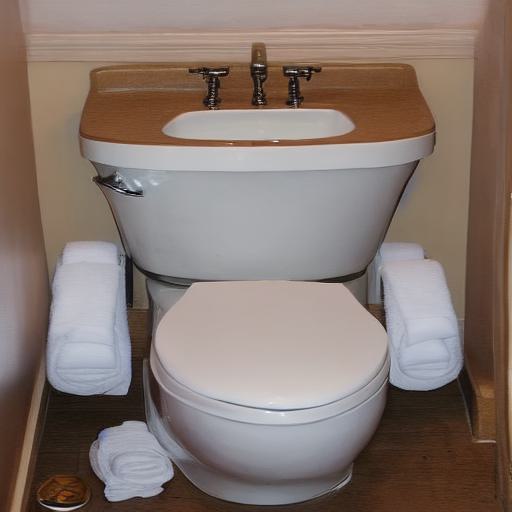}\\
    \includegraphics[width=1\textwidth]{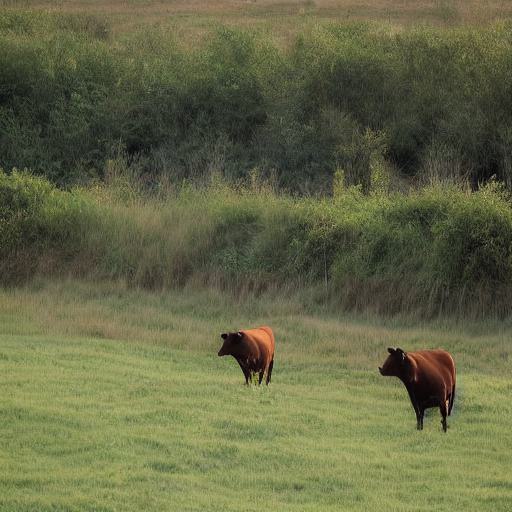}\\
    \includegraphics[width=1\textwidth]{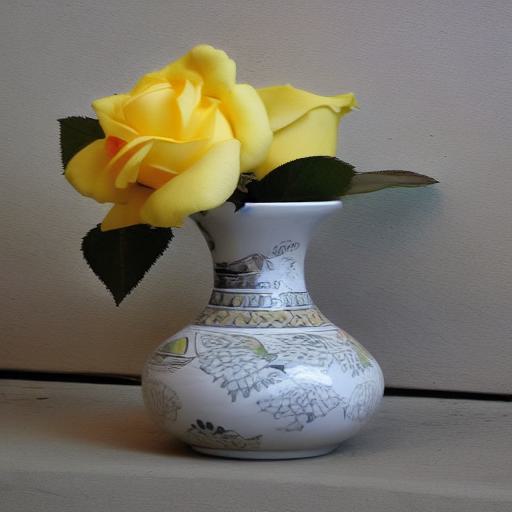}\\
    \includegraphics[width=1\textwidth]{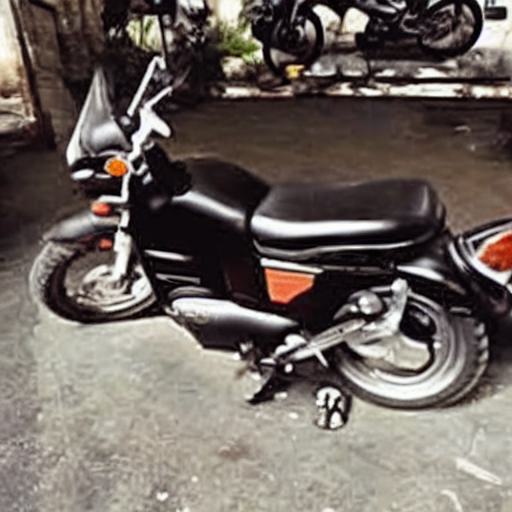}
\end{minipage}
\begin{minipage}[c]{0.16\textwidth}
    \makebox[1\textwidth]{RepControlNet-} \\
    \makebox[1\textwidth]{(SD1.5)} \\
    \includegraphics[width=1\textwidth]{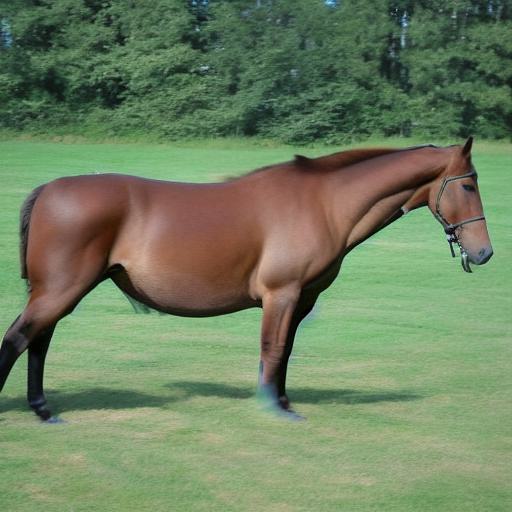}\\
    \includegraphics[width=1\textwidth]{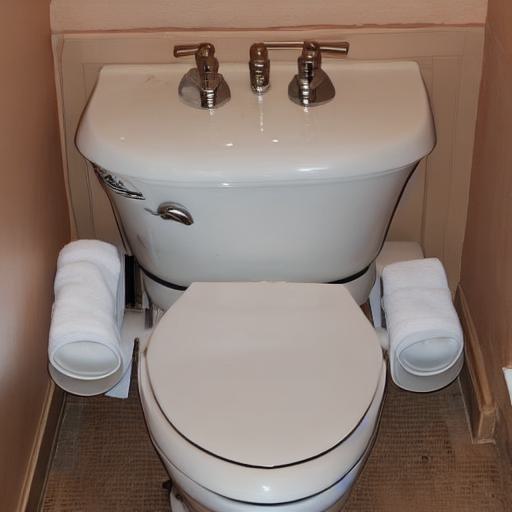}\\
    \includegraphics[width=1\textwidth]{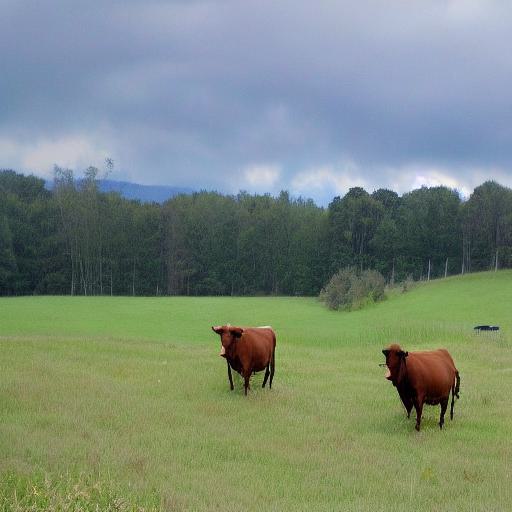}\\
    \includegraphics[width=1\textwidth]{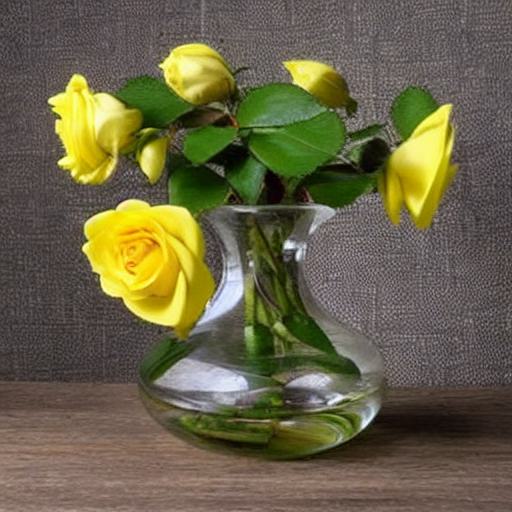}\\
    \includegraphics[width=1\textwidth]{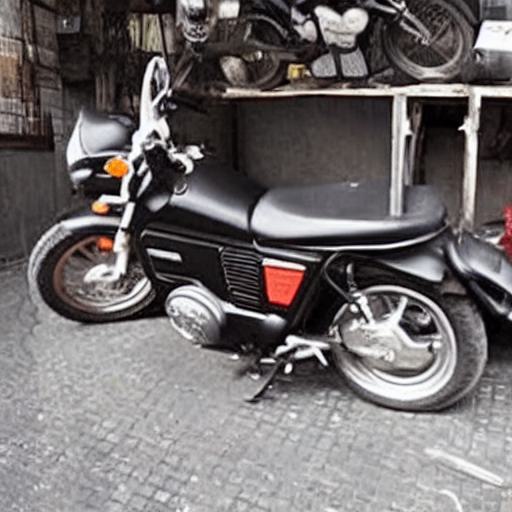}
\end{minipage}
% \begin{minipage}[c]{0.16\textwidth}
%     \makebox[1\textwidth]{RepControlNet+} \\
%     \makebox[1\textwidth]{(SDXL)} \\
%     \includegraphics[width=1\textwidth]{figs/SS_COCO_10000/ss_xl_coco10000/samples/8975.jpg}\\
%     \includegraphics[width=1\textwidth]{figs/SS_COCO_10000/ss_xl_coco10000/samples/540.jpg}\\
%     \includegraphics[width=1\textwidth]{figs/SS_COCO_10000/ss_xl_coco10000/samples/843.jpg}\\
%     \includegraphics[width=1\textwidth]{figs/SS_COCO_10000/ss_xl_coco10000/samples/19468.jpg}\\
%     \includegraphics[width=1\textwidth]{figs/SS_COCO_10000/ss_xl_coco10000/samples/38.jpg}
% \end{minipage}
\begin{minipage}[c]{0.16\textwidth}
    \makebox[1\textwidth]{RepControlNet-} \\
    \makebox[1\textwidth]{(SDXL)} \\
    \includegraphics[width=1\textwidth]{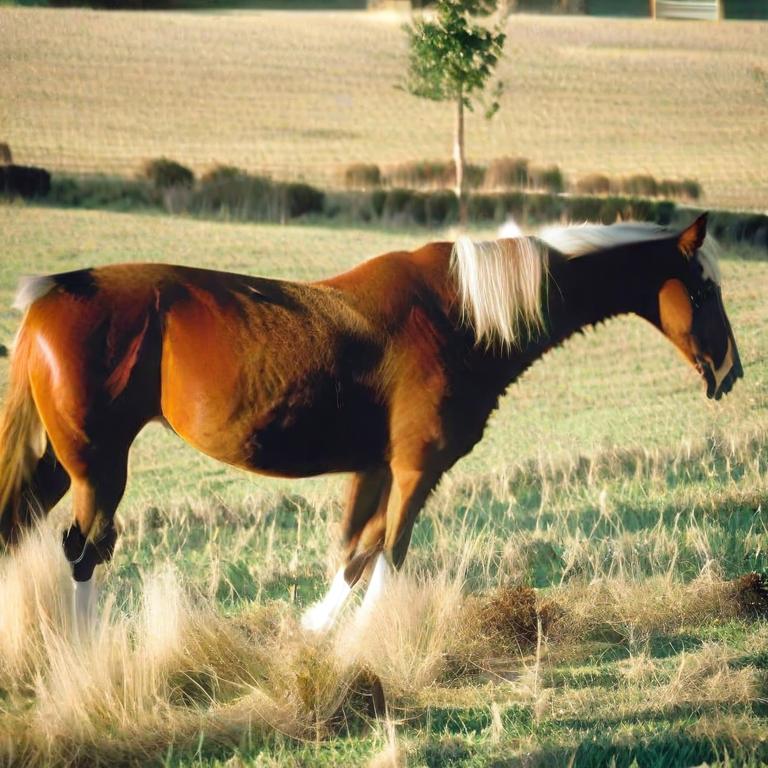}\\
    \includegraphics[width=1\textwidth]{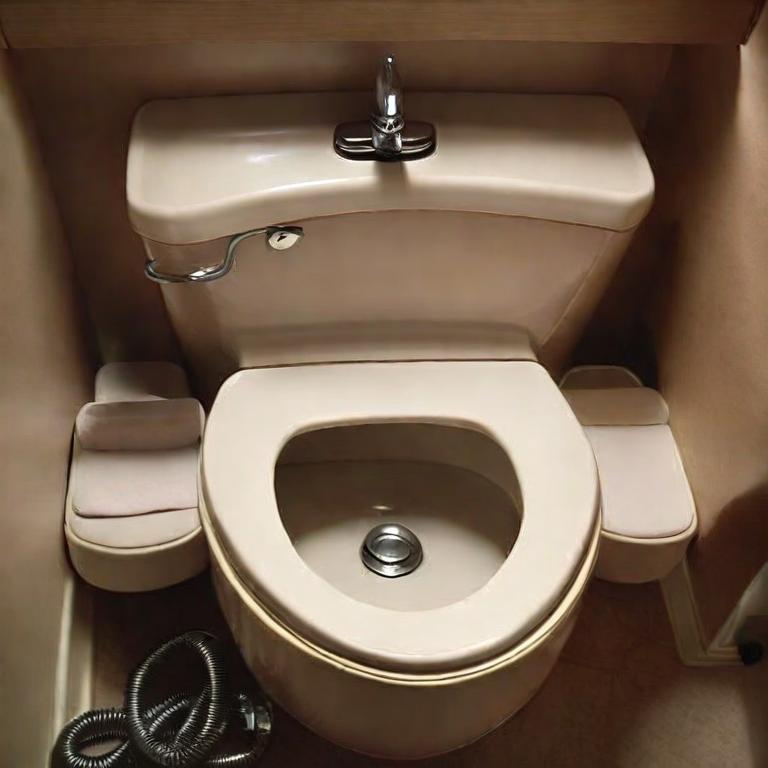}\\
    \includegraphics[width=1\textwidth]{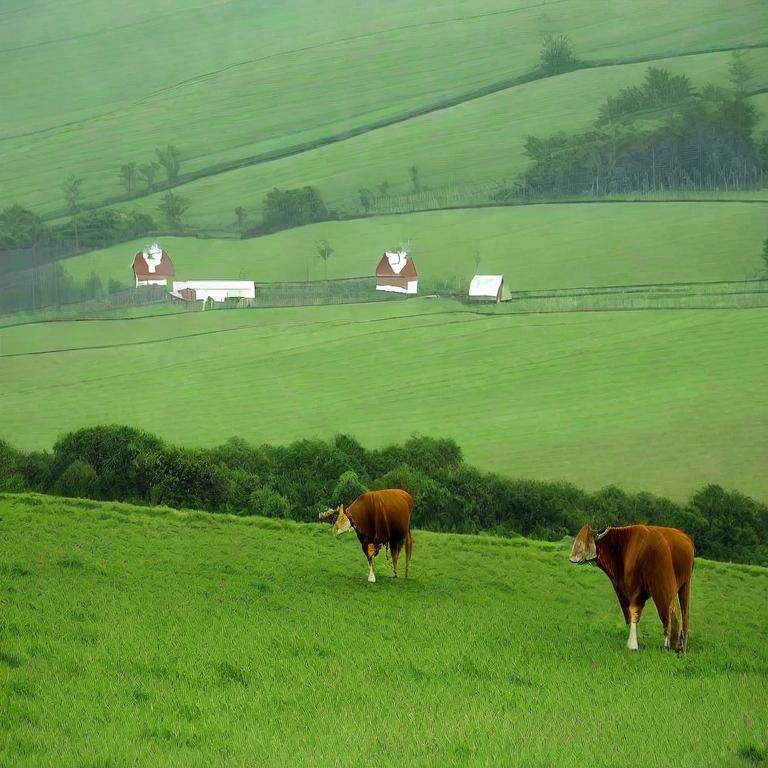}\\
    \includegraphics[width=1\textwidth]{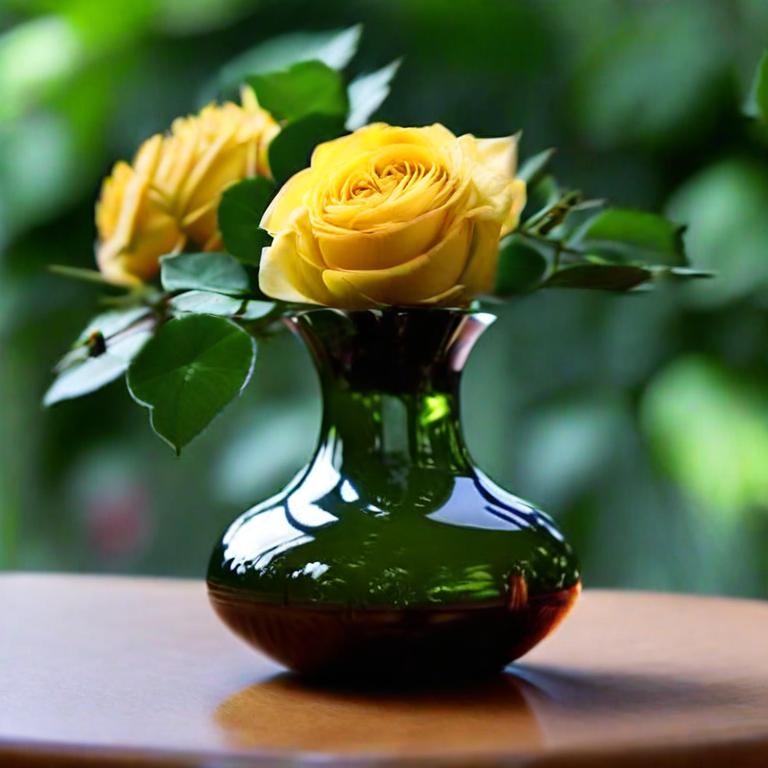}\\
    \includegraphics[width=1\textwidth]{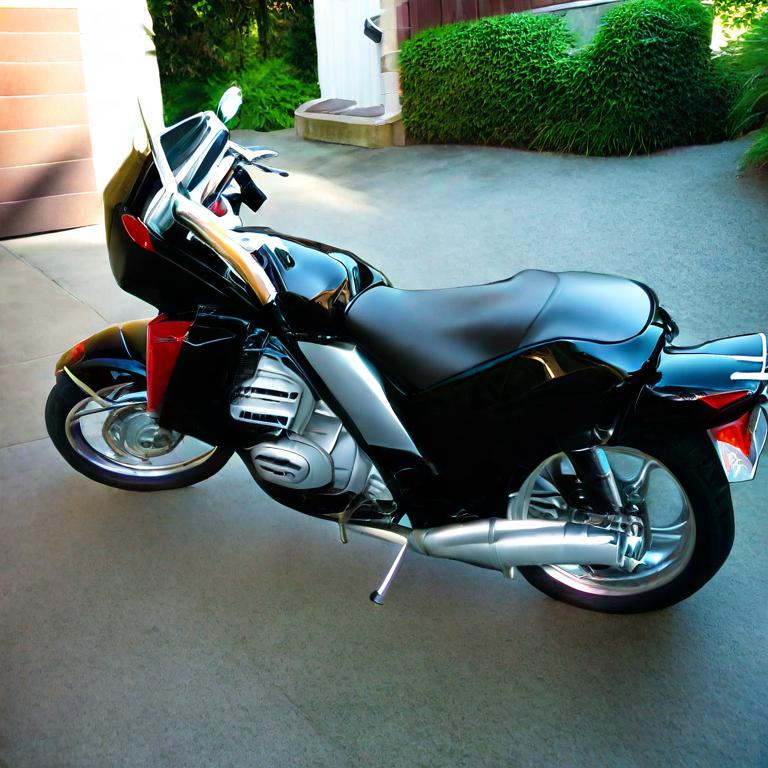}
\end{minipage}
\caption{Image generation results conditioned by semantic segmentation compared with baseline ControlNet. Captions: 
A brown horse with a white mane is on the grass.
A tan toilet and sink combination in a small room.
% A cat sitting by a laptop but staring at a person.
Two brown cows on a pasture of green grass.
A beautiful yellow rose is seen in a small vase.
A black Honda motorcycle parked in front of a garage.}
\label{fig:ss_compare}
\end{figure*}

\begin{figure*}[!t]
\begin{center}
\includegraphics[width=0.92\textwidth]{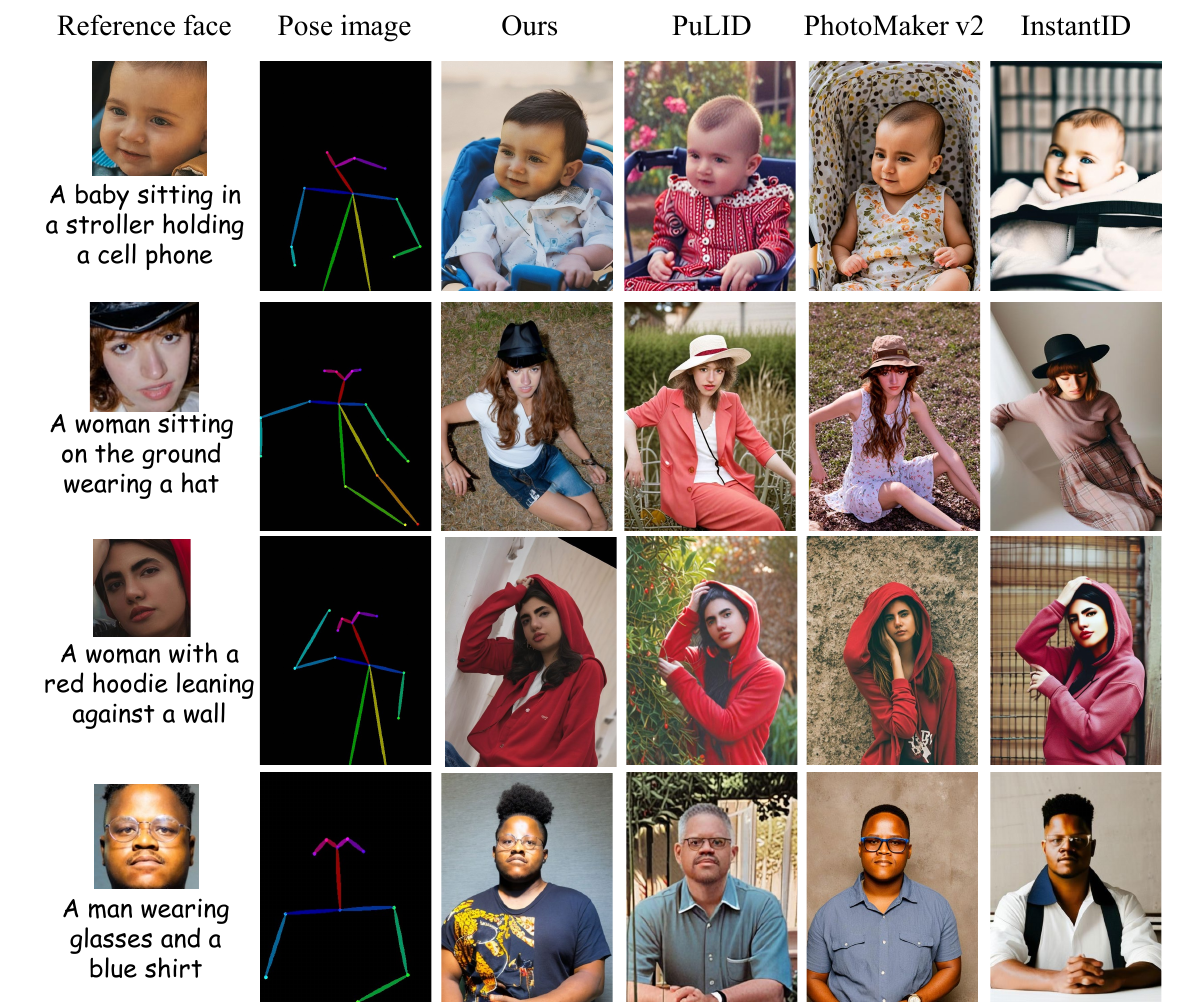}
\end{center}
\caption{Skeleton guided portrait fidelity generation qualitative comparison. Images are collected from Unsplash-50~\cite{gal2024lcm} dataset.}
\label{fig:face}
\end{figure*}

\section{Experiments}

To validate the effectiveness and efficiency of our proposed RepControlNet, we conduct qualitative experiments for image generation with various conditions and compare it with baseline model conditioned by semantic segmentation maps. Furthermore, we perform quantitative evaluation experiments, including FID and CLIP-score for conditional fidelity, as well as FLOPs, MACs and the number of parameters for performance.

\subsection{Implementation details}

We implement RepControlNet on both SD1.5 \cite{Rombach_2022_CVPR} and SDXL \cite{podell2023sdxl} to test various conditions, including Canny Edge \cite{canny1986computational}, Depth Map \cite{ranftl2020towards}, Semantic Segmentation \cite{zhou2019semantic,caesar2018cvpr}.
% and  Openpose\cite{jiang2023rtmpose}.

In the following sections, RepControlNet+ refers to RepControlNet during training without reparameterization, and RepControlNet- refers to RepControlNet during inference with reparameterization.

\subsection{Qualitative comparison}

\subsubsection {Canny Edge.} We take RGB images and captions from  the LLAVA-558k \cite{liu2024visual} dataset and generate edge images with a Canny edge detector \cite{canny1986computational} (with random thresholds) to obtain 558K edge-image-caption pairs. The base model used is SD1.5. Figure \ref{fig:canny} shows the generated images and the corresponding input edge-image-caption pairs.

\subsubsection {Depth Map.} We take RGB images and Depth maps from the outdoor portion of DIODE dataset \cite{vasiljevic2019diode} and generate captions for RGB images by BLIP2 \cite{li2023blip} to obtain 16K depth-image-caption pairs. The base model used is SD1.5. Figure \ref{fig:depth} shows the generated images and the corresponding input depth-image-caption pairs.

\subsubsection {Semantic Segmentation.} 
We take RGB images, semantic segmentation maps and captions from COCO-Stuff dataset \cite{caesar2018coco} to obtain segmentation-image-caption pairs, including 118K pairs for training and 5K for testing. The base model used is SD1.5. Figure \ref{fig:ss_coco} shows the generated images and the corresponding input segmentation-image-caption pairs.
With the ADE20K dataset \cite{zhou2017scene} captioned by BLIP2, we obtain 27K segmentation-image-caption pairs, including 25K pairs for the training and 2K for testing. The base model used is SD1.5 and SDXL. Figure \ref{fig:ss_ade} shows the generated images and the corresponding input segmentation-image-caption pairs.

% \subsubsection {Openpose.} 

% \begin{figure}[ht]
% \centering
% \begin{minipage}[c]{1\columnwidth}
%     \makebox[0.32\columnwidth]{Human pose}
%     \makebox[0.32\columnwidth]{Original RGB}
%     \makebox[0.32\columnwidth]{Generated}
% \end{minipage}
% \\
% \includegraphics[width=0.32\columnwidth]{figs/SS_COCO_10000/control/validation-ADE_val_00000001.png}
% \includegraphics[width=0.32\columnwidth]{figs/SS_COCO_10000/samples/validation-ADE_val_00000001.jpg}
% \includegraphics[width=0.32\columnwidth]{figs/SS_COCO_10000/reconstruction/validation-ADE_val_00000001.jpg}
% \\
% \includegraphics[width=0.32\columnwidth]{figs/SS_COCO_10000/control/validation-ADE_val_00000049.png}
% \includegraphics[width=0.32\columnwidth]{figs/SS_COCO_10000/samples/validation-ADE_val_00000049.jpg}
% \includegraphics[width=0.32\columnwidth]{figs/SS_COCO_10000/reconstruction/validation-ADE_val_00000049.jpg}
% \\
% \includegraphics[width=0.32\columnwidth]{figs/SS_COCO_10000/control/validation-ADE_val_00001695.png}
% \includegraphics[width=0.32\columnwidth]{figs/SS_COCO_10000/samples/validation-ADE_val_00001695.jpg}
% \includegraphics[width=0.32\columnwidth]{figs/SS_COCO_10000/reconstruction/validation-ADE_val_00001695.jpg}
% \caption{Image generation results with Openpose conditioning.}
% \label{fig:ss_pose}
% \end{figure}

\subsubsection {Compare with baseline.} 

We further compare RepControlNet with baseline model ControlNet conditioned by semantic segmentation maps with random samples from COCO-Stuff validation set, as shown in Figure \ref{fig:ss_compare}. It can be seen that RepControlNet can achieve comparable controllable generation effects to ControlNet, and the generation results by RepControlNet- have only minor differences compared to those by RepControlNet+.
\subsubsection {Skeleton.} 
% 为了验证Rep的通用性，我们在method中介绍了将其结合到自研的dentity-Preserving generation算法中进行实验。对比实验结果如图7所示，训练过程使用skeleton作为Rep的condition。结果表明其可以方便且有效地移植到别的算法方案中。
To verify the generality of RepControlNet, we introduced it into our self-developed Identity-Preserving generation algorithm in the methods section for experiments. The comparison of experimental results is shown in Figure~\ref{fig:face}, where the training process uses the skeleton as the condition for RepControlNet. The results indicate that it can be easily and effectively transplanted into other algorithm schemes.

\subsection{Quantitative comparison}

To quantitatively evaluate the conditioning fidelity, we use the ADE20K dataset for training and the COCO-Stuff dataset for testing similar to ControlNet \cite{zhang2023adding}. For a comprehensive evaluation, we use FID score ~\cite{heusel2017gans} and text-image CLIP scores \cite{radford2021learning} as metrics. To be specific, we select 10000 random prompts from the COCO validation set and use OneFormer to generate paired semantic segmentation maps of these samples. 
For SD1.5 base model, we use random crops of $512\times512$ images as FID’s target set and $512\times512$ images generated by different methods using the paired semantic segmentation maps as the FID’s source set. For SDXL base model, we use images of the resolution $768\times768$.
% We use random crops of $512\times512$ images as FID’s target set and $512\times512$ images generated by different methods using the paired semantic segmentation maps as the FID’s source set.
The results are presented in Table \ref{table:semantic_segment}. It can be seen that our method can achieve comparable or even superior performance to methods like ControlNet that use additional parameters, without increasing the number of parameters.

\begin{table}[!t]\centering
    \begin{tabular}{c|cc}
        \hline
        Method & FID$\downarrow$ & CLIP-score$\uparrow$ \\ 
        \hline
        VQGAN (seg.) & 26.28 & 0.17\\
        LDM (seg.) & 25.35 & 0.18\\
        PIPT (seg.) & 19.74 & 0.20\\
        ControlNet-lite(SD1.5) & 17.92 & 0.26\\
        ControlNet(SD1.5) & 15.27 & 0.26\\
        RepControlNet+(SD1.5) & \textbf{14.79}  & \textbf{0.27}\\
        RepControlNet-(SD1.5) & 14.80 & 0.27\\
        \hline
        ControlNet++(SDXL) & 27.04 & 0.24 \\
        RepControlNet+(SDXL) &  \textbf{19.08}  & \textbf{0.25} \\
        RepControlNet-(SDXL) & \textbf{19.08}  & \textbf{0.25} \\
        \hline
    \end{tabular}
    \caption{Quantitative Evaluation for image generation conditioned by semantic segmentation. We report FID and CLIP text-image score for our method and other baselines, including VQGAN \cite{esser2021taming}, LDM \cite{rombach2022high}, PIPT \cite{wang2022pretraining}, ControlNet and ControlNet++\cite{controlnet++}. RepControlNet+ and RepControlNet- represent before and after reparameterization separately.}
    \label{table:semantic_segment}
\end{table}

To validate the advantages of RepControlNet over ControlNet in reducing the parameters and computations, we also calculate the number of parameters and test the speed of every model with a batch size of 1 for a single step. Here, we use FLOPs (floating-point operations) and MACs (multiply-accumulate operations) as metrics for computational speed, as shown in Table \ref{table:compute}. It can be seen that RepControlNet adds only a negligible amount of computational overhead and model parameters during the inference process compared to SD base models, which has significant advantages compared to ControlNet.

\begin{table}[!t]\centering
    \begin{tabular}{c|ccc}
        \hline
        Method & FLOPs & MACs & Param. \\ 
        \hline
        SD1.5  & 0.68 T & 0.34 T  & 1066 M \\
        ControlNet(SD1.5) & 0.91 T  & 0.46 T & 1427 M \\
        RepControlNet+(SD1.5) & 1.37 T & 0.68 T & 1925 M \\
        RepControlNet-(SD1.5) & \textbf{0.69 T} & \textbf{0.35 T} & \textbf{1067 M} \\
        \hline
        SDXL & 6.78 T & 3.39 T  & 3468 M \\
        RepControlNet+(SDXL) & 13.58 T & 6.79 T  & 6030 M \\
        RepControlNet-(SDXL) & \textbf{6.81 T} & \textbf{3.40 T} & \textbf{3470 M} \\
        \hline
    \end{tabular}
    \caption{Comparison of FLOPs, MACs and parameters between SD1.5, SDXL, ControlNet and RepcontrolNet.}
    \label{table:compute}
\end{table}

% \subsection{Ablation study}
%加不加Reparam可视化对比
%调整base loss权重对比

\begin{table}[!t]
% \vspace{-1.0em}
\centering
  \setlength{\tabcolsep}{0pt} 
  \begin{tabular*}{\columnwidth}{@{\extracolsep{\fill}}c|cccc@{}}
    \hline
    \textbf{Method}   & CLIP-score$\uparrow$ &  FaceSim$\uparrow$   & FID$\downarrow$ \\
    \hline
    % IP-Adapter~\cite{blattmann2023stable} &  w/o & 88.1 & 19.5 &40.3 &86.5 & \textbf{190.5}\\
    IP-Adapter  & 0.20 &41.1 & 213.8\\
    % InstantID~\cite{blattmann2023stable} &  w/o &87.4 & 20.1  & \textbf{68.6}  & 73.1 &252.0\\
    InstantID   & 0.19  &57.7 &220.3\\
    % PhotomakeV2~\cite{zhang2023pia}  &  w/o &88.0  & \textbf{21.4} &  50.5 &    80.3 &220.1 \\
    PhotomakeV2  & \textbf{0.21}  &  50.5 &  216.1 \\
    % PULID-512~\cite{blattmann2023stable} &  w/o &0.8608 & 17.9  & 0.4923  &  0.8198 &\\
    % PuLID~\cite{blattmann2023stable} &  w/o &88.1 & 20.9  & 61.0  & 86.5 &  240.4\\
    PuLID& 0.19 & 47.0  & 229.9\\
    % \textbf{MagicID}~\cite{zhang2023pia}  &  w/o &90.0  & 20.9 & 68.5 &  82.4 &214.3   \\
    RepControlNet- &  \textbf{0.21}  & \textbf{60.7}  &\textbf{195.8} \\
    \hline
  \end{tabular*}
 \caption{Skeleton guided portrait fidelity generation quantitative comparison. We use pretrained skeleton guided RepControlNet. The best average performance is in bold. ↑ indicates higher metric value and represents better performance and vice versa.}
\label{tab:Quantitative_halfbody}
% \vspace{-2.0em}
\end{table}
We further conduct Identity-Preserving generation quantitative experiments to verify the portability of our RepControlNet. We evaluated the identity-preserving generation capabilities of various models on the Unsplash-50~\cite{gal2024lcm} dataset by comprehensively assessing their performance through metrics such as CLIP scores, facial similarity(FaceSim), and FID. We benchmarked against sota open-source solutions, including InstantID~\cite{wang2024instantid}, PuLID ~\cite{guo2024pulid}, and PhotoMakerV2~\cite{li2023photomaker}. To ensure a fair comparison, we adhered to their respective official configurations by incorporating default prompts to enhance generation quality and utilized the Openpose ControlNet model to constrain poses during the generation process. On the unsplash test set, as listed in Table.~\ref{tab:Quantitative_halfbody}, our method achieved the best performance across all metrics, with a significant lead in the FaceSim metric. These results fully validate the portability of our proposed RepControlNet from objective metrics.

\section{Conclusions}
% 我们提出了一种不需要额外运算量的条件控制生成方法，RepControlNet.在SD1.5和SDXL两个基座模型，Canny、Depthmap、semantic map、skeleton四种模态上进行了充分实验。为验证RepControlNet的通用性，我们将其结合进自研的dentity-Preserving generation method，以skeleton作为控制条件进行训练，实验的主客观指标验证了其可移植性和高效性。
 We propose a conditional control generation method that does not require additional computational overhead, named RepControlNet. Extensive experiments were conducted on two base models, SD1.5 and SDXL, across four modalities: Canny, Depthmap, semantic map, and skeleton. To verify the generality of RepControlNet, we integrated it into our self-developed Identity-Preserving generation method, using the skeleton as the control condition for training. The experimental results, both subjective and objective metrics, validated its portability and efficiency. There is still a long way to go to popularize the application of diffusion models, and we hope that our work can bring some inspiration to researchers.

% Generated by IEEEtran.bst, version: 1.14 (2015/08/26)


\begin{thebibliography}{10}
\providecommand{\url}[1]{#1}
\csname url@samestyle\endcsname
\providecommand{\newblock}{\relax}
\providecommand{\bibinfo}[2]{#2}
\providecommand{\BIBentrySTDinterwordspacing}{\spaceskip=0pt\relax}
\providecommand{\BIBentryALTinterwordstretchfactor}{4}
\providecommand{\BIBentryALTinterwordspacing}{\spaceskip=\fontdimen2\font plus
\BIBentryALTinterwordstretchfactor\fontdimen3\font minus \fontdimen4\font\relax}
\providecommand{\BIBforeignlanguage}[2]{{%
\expandafter\ifx\csname l@#1\endcsname\relax
\typeout{** WARNING: IEEEtran.bst: No hyphenation pattern has been}%
\typeout{** loaded for the language `#1'. Using the pattern for}%
\typeout{** the default language instead.}%
\else
\language=\csname l@#1\endcsname
\fi
#2}}
\providecommand{\BIBdecl}{\relax}
\BIBdecl

\bibitem{li2024snapfusion}
Y.~Li, H.~Wang, Q.~Jin, J.~Hu, P.~Chemerys, Y.~Fu, Y.~Wang, S.~Tulyakov, and J.~Ren, ``Snapfusion: Text-to-image diffusion model on mobile devices within two seconds,'' \emph{Advances in Neural Information Processing Systems}, vol.~36, 2024.

\bibitem{xu2024ufogen}
Y.~Xu, Y.~Zhao, Z.~Xiao, and T.~Hou, ``Ufogen: You forward once large scale text-to-image generation via diffusion gans,'' in \emph{Proceedings of the IEEE/CVF Conference on Computer Vision and Pattern Recognition}, 2024, pp. 8196--8206.

\bibitem{ding2021repvgg}
X.~Ding, X.~Zhang, N.~Ma, J.~Han, G.~Ding, and J.~Sun, ``Repvgg: Making vgg-style convnets great again,'' in \emph{Proceedings of the IEEE/CVF conference on computer vision and pattern recognition}, 2021, pp. 13\,733--13\,742.

\bibitem{Wang_2024_CVPR}
A.~Wang, H.~Chen, Z.~Lin, J.~Han, and G.~Ding, ``Repvit: Revisiting mobile cnn from vit perspective,'' in \emph{Proceedings of the IEEE/CVF Conference on Computer Vision and Pattern Recognition (CVPR)}, June 2024, pp. 15\,909--15\,920.

\bibitem{song2023consistency}
Y.~Song, P.~Dhariwal, M.~Chen, and I.~Sutskever, ``Consistency models,'' \emph{arXiv preprint arXiv:2303.01469}, 2023.

\bibitem{salimans2022progressive}
T.~Salimans and J.~Ho, ``Progressive distillation for fast sampling of diffusion models,'' \emph{arXiv preprint arXiv:2202.00512}, 2022.

\bibitem{canny1986computational}
J.~Canny, ``A computational approach to edge detection,'' \emph{IEEE Transactions on pattern analysis and machine intelligence}, pp. 679--698, 1986.

\bibitem{Rombach_2022_CVPR}
R.~Rombach, A.~Blattmann, D.~Lorenz, P.~Esser, and B.~Ommer, ``High-resolution image synthesis with latent diffusion models,'' in \emph{Proceedings of the IEEE/CVF Conference on Computer Vision and Pattern Recognition (CVPR)}, June 2022, pp. 10\,684--10\,695.

\bibitem{podell2023sdxl}
D.~Podell, Z.~English, K.~Lacey, A.~Blattmann, T.~Dockhorn, J.~M{\"u}ller, J.~Penna, and R.~Rombach, ``Sdxl: Improving latent diffusion models for high-resolution image synthesis,'' \emph{arXiv preprint arXiv:2307.01952}, 2023.

\bibitem{ranftl2020towards}
R.~Ranftl, K.~Lasinger, D.~Hafner, K.~Schindler, and V.~Koltun, ``Towards robust monocular depth estimation: Mixing datasets for zero-shot cross-dataset transfer,'' \emph{IEEE transactions on pattern analysis and machine intelligence}, vol.~44, no.~3, pp. 1623--1637, 2020.

\bibitem{jiang2023rtmpose}
T.~Jiang, P.~Lu, L.~Zhang, N.~Ma, R.~Han, C.~Lyu, Y.~Li, and K.~Chen, ``Rtmpose: Real-time multi-person pose estimation based on mmpose,'' \emph{arXiv preprint arXiv:2303.07399}, 2023.

\bibitem{zhou2019semantic}
B.~Zhou, H.~Zhao, X.~Puig, T.~Xiao, S.~Fidler, A.~Barriuso, and A.~Torralba, ``Semantic understanding of scenes through the ade20k dataset,'' \emph{International Journal of Computer Vision}, vol. 127, no.~3, pp. 302--321, 2019.

\bibitem{caesar2018cvpr}
H.~Caesar, J.~Uijlings, and V.~Ferrari, ``Coco-stuff: Thing and stuff classes in context,'' in \emph{Computer vision and pattern recognition (CVPR), 2018 IEEE conference on}.\hskip 1em plus 0.5em minus 0.4em\relax IEEE, 2018.

\bibitem{liu2024visual}
H.~Liu, C.~Li, Q.~Wu, and Y.~J. Lee, ``Visual instruction tuning,'' \emph{Advances in neural information processing systems}, vol.~36, 2024.

\bibitem{vasiljevic2019diode}
I.~Vasiljevic, N.~Kolkin, S.~Zhang, R.~Luo, H.~Wang, F.~Z. Dai, A.~F. Daniele, M.~Mostajabi, S.~Basart, M.~R. Walter \emph{et~al.}, ``Diode: A dense indoor and outdoor depth dataset,'' \emph{arXiv preprint arXiv:1908.00463}, 2019.

\bibitem{li2023blip}
J.~Li, D.~Li, S.~Savarese, and S.~Hoi, ``Blip-2: Bootstrapping language-image pre-training with frozen image encoders and large language models,'' in \emph{International conference on machine learning}.\hskip 1em plus 0.5em minus 0.4em\relax PMLR, 2023, pp. 19\,730--19\,742.

\bibitem{caesar2018coco}
H.~Caesar, J.~Uijlings, and V.~Ferrari, ``Coco-stuff: Thing and stuff classes in context,'' in \emph{Proceedings of the IEEE conference on computer vision and pattern recognition}, 2018, pp. 1209--1218.

\bibitem{zhou2017scene}
B.~Zhou, H.~Zhao, X.~Puig, S.~Fidler, A.~Barriuso, and A.~Torralba, ``Scene parsing through ade20k dataset,'' in \emph{Proceedings of the IEEE conference on computer vision and pattern recognition}, 2017, pp. 633--641.

\bibitem{zhang2023adding}
L.~Zhang, A.~Rao, and M.~Agrawala, ``Adding conditional control to text-to-image diffusion models,'' in \emph{Proceedings of the IEEE/CVF International Conference on Computer Vision}, 2023, pp. 3836--3847.

\bibitem{mou2024t2i}
C.~Mou, X.~Wang, L.~Xie, Y.~Wu, J.~Zhang, Z.~Qi, and Y.~Shan, ``T2i-adapter: Learning adapters to dig out more controllable ability for text-to-image diffusion models,'' in \emph{Proceedings of the AAAI Conference on Artificial Intelligence}, vol.~38, 2024, pp. 4296--4304.

\bibitem{ye2023ip}
H.~Ye, J.~Zhang, S.~Liu, X.~Han, and W.~Yang, ``Ip-adapter: Text compatible image prompt adapter for text-to-image diffusion models,'' \emph{arXiv preprint arXiv:2308.06721}, 2023.

\bibitem{wang2024instantid}
Q.~Wang, X.~Bai, H.~Wang, Z.~Qin, and A.~Chen, ``Instantid: Zero-shot identity-preserving generation in seconds,'' \emph{arXiv preprint arXiv:2401.07519}, 2024.

\bibitem{radford2021learning}
A.~Radford, J.~W. Kim, C.~Hallacy, A.~Ramesh, G.~Goh, S.~Agarwal, G.~Sastry, A.~Askell, P.~Mishkin, J.~Clark \emph{et~al.}, ``Learning transferable visual models from natural language supervision,'' in \emph{International conference on machine learning}.\hskip 1em plus 0.5em minus 0.4em\relax PMLR, 2021, pp. 8748--8763.

\bibitem{schuhmann2022laion}
C.~Schuhmann, R.~Beaumont, R.~Vencu, C.~Gordon, R.~Wightman, M.~Cherti, T.~Coombes, A.~Katta, C.~Mullis, M.~Wortsman \emph{et~al.}, ``Laion-5b: An open large-scale dataset for training next generation image-text models,'' \emph{Advances in Neural Information Processing Systems}, vol.~35, pp. 25\,278--25\,294, 2022.

\bibitem{esser2021taming}
P.~Esser, R.~Rombach, and B.~Ommer, ``Taming transformers for high-resolution image synthesis,'' in \emph{Proceedings of the IEEE/CVF conference on computer vision and pattern recognition}, 2021, pp. 12\,873--12\,883.

\bibitem{rombach2022high}
R.~Rombach, A.~Blattmann, D.~Lorenz, P.~Esser, and B.~Ommer, ``High-resolution image synthesis with latent diffusion models,'' in \emph{Proceedings of the IEEE/CVF conference on computer vision and pattern recognition}, 2022, pp. 10\,684--10\,695.

\bibitem{wang2022pretraining}
T.~Wang, T.~Zhang, B.~Zhang, H.~Ouyang, D.~Chen, Q.~Chen, and F.~Wen, ``Pretraining is all you need for image-to-image translation,'' \emph{arXiv preprint arXiv:2205.12952}, 2022.

\bibitem{gal2024lcm}
R.~Gal, O.~Lichter, E.~Richardson, O.~Patashnik, A.~H. Bermano, G.~Chechik, and D.~Cohen-Or, ``Lcm-lookahead for encoder-based text-to-image personalization,'' \emph{arXiv preprint arXiv:2404.03620}, 2024.

\bibitem{guo2024pulid}
Z.~Guo, Y.~Wu, Z.~Chen, L.~Chen, and Q.~He, ``Pulid: Pure and lightning id customization via contrastive alignment,'' \emph{arXiv preprint arXiv:2404.16022}, 2024.

\bibitem{li2023photomaker}
Z.~Li, M.~Cao, X.~Wang, Z.~Qi, M.-M. Cheng, and Y.~Shan, ``Photomaker: Customizing realistic human photos via stacked id embedding,'' in \emph{IEEE Conference on Computer Vision and Pattern Recognition (CVPR)}, 2024.

\bibitem{controlnet++}
D.~Lonarkar, ``Controlnet++: All-in-one controlnet for image generations and editing,'' \url{https://github.com/xinsir6/ControlNetPlus}, 2024.

\bibitem{DBLP:journals/corr/abs-2103-00020}
A.~Radford, J.~W. Kim, C.~Hallacy, A.~Ramesh, G.~Goh, S.~Agarwal, G.~Sastry, A.~Askell, P.~Mishkin, J.~Clark, G.~Krueger, and I.~Sutskever, ``Learning transferable visual models from natural language supervision,'' \emph{CoRR}, 2021.

\bibitem{deng2018arcface}
J.~Deng, J.~Guo, X.~Niannan, and S.~Zafeiriou, ``Arcface: Additive angular margin loss for deep face recognition,'' in \emph{CVPR}, 2019.

\bibitem{heusel2017gans}
M.~Heusel, H.~Ramsauer, T.~Unterthiner, B.~Nessler, and S.~Hochreiter, ``Gans trained by a two time-scale update rule converge to a local nash equilibrium,'' \emph{Advances in neural information processing systems}, vol.~30, 2017.

\end{thebibliography}
\end{document}